\documentclass[10pt,twocolumn,letterpaper]{article}

\usepackage{wacv} %

\usepackage{multirow}
\usepackage{makecell}
\usepackage{tikz}
\usepackage{fontawesome5}
\usepackage{mydefs}

\definecolor{wacvblue}{rgb}{0.21,0.49,0.74}
\usepackage[pagebackref,breaklinks,colorlinks,allcolors=wacvblue]{hyperref}

\def\wacvPaperID{637} %
\def\confName{WACV}
\def\confYear{2026}

\title{Online Episodic Memory Visual Query Localization \\ with Egocentric Streaming Object Memory}

\author{\vspace{.15cm} Zaira Manigrasso$^{1\ast}$
\quad\ Matteo Dunnhofer$^{1,3\ast}$
\quad\ Antonino Furnari$^{2\ast}$ \\
\vspace{.15cm}
Moritz Nottebaum$^{1}$
\quad\ Antonio Finocchiaro$^{2}$
\quad\ Davide Marana$^{1}$ 
\quad\ Rosario Forte$^{2}$ \\
Giovanni Maria Farinella$^{2\star}$ 
\quad\ Christian Micheloni$^{1\star}$
\\
\\
$^{1}$ University of Udine, Italy \quad
$^{2}$ University of Catania, Italy \quad
$^{3}$ York University, Canada \\
}

\begin{document}
\maketitle
{\let\thefootnote\relax\footnote{$^{\ast}$ denotes equal contribution. $^{\star}$ denotes equal supervision.}

\begin{abstract}
Episodic memory retrieval enables wearable cameras to recall objects or events previously observed in video.
However, existing formulations assume an ``offline'' setting with full video access at query time, limiting their applicability in real-world scenarios with power and storage-constrained wearable devices.
Towards more application-ready episodic memory systems, we introduce \textit{Online Visual Query 2D} (OVQ2D), a task where models process video streams online, observing each frame only once, and retrieve object localizations using a compact memory instead of full video history.
We address OVQ2D with ESOM (Egocentric Streaming Object Memory), a novel framework integrating an \textit{object discovery module}, an \textit{object tracking module}, and a \textit{memory module} that find, track, and store spatio-temporal object information for efficient querying.
Experiments on Ego4D demonstrate ESOM's superiority over other online approaches, though OVQ2D remains challenging, with top performance at only ~4\% success. ESOM’s accuracy increases markedly with perfect object tracking (31.91\%), discovery (40.55\%), or both (81.92\%), underscoring the need of applied research on these components.
\end{abstract}

\begin{figure}[t]
    \centering
    \includegraphics[width=1\columnwidth]{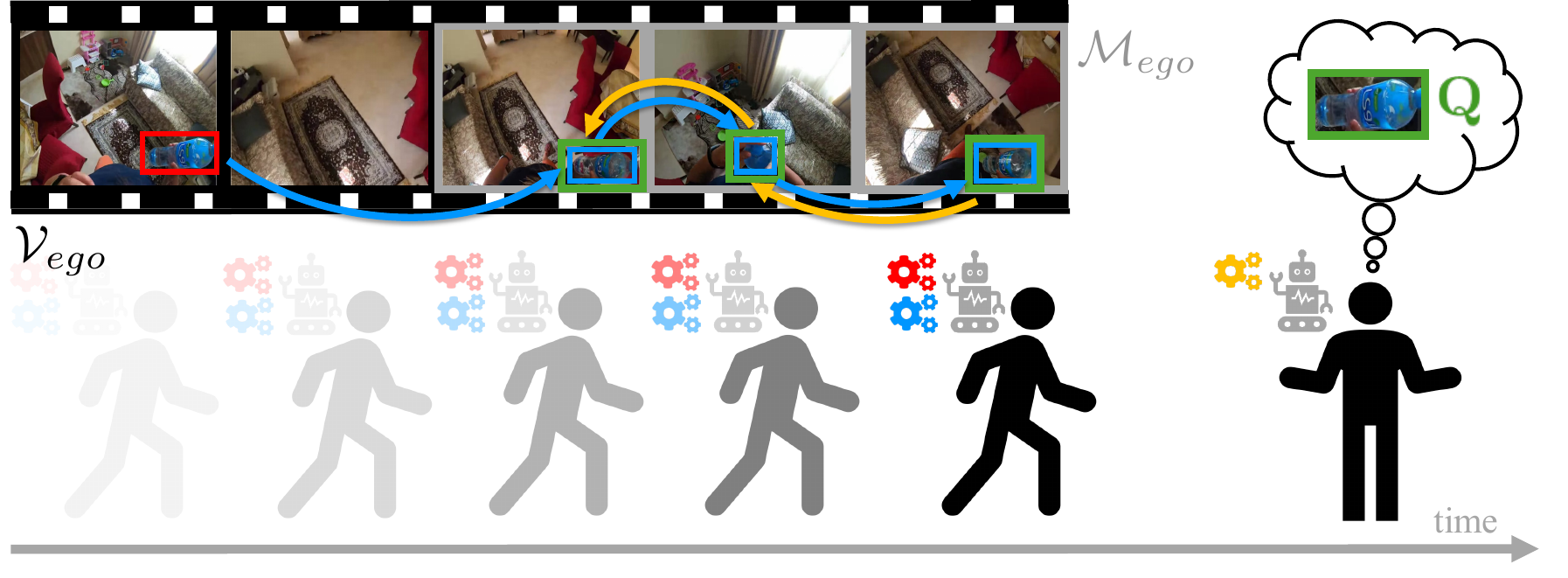}
    \caption{\textbf{Online visual query localization via object memorization and retrieval.}
    We tackle the problem of online episodic memory and propose ESOM (\img{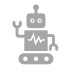}), an architecture that processes  an egocentric video ($\video$) online and only once to detect (\textcolor[HTML]{FF2140}{\faCogs}), track (\textcolor[HTML]{0F80FE}{\faCogs}), and memorize (\textcolor[HTML]{7F7F7F}{$\memory$}) user-relevant objects and frames. When retrieving a visual query (\textcolor[HTML]{4DA72E}{$\query$}), ESOM (\img{images/robot.jpg}) searches (\textcolor[HTML]{FFC000}{\faCogs}) its memory (\textcolor[HTML]{7F7F7F}{$\memory$}) for the most recent instance where the query was spatio-temporally localized. By avoiding video storage, ESOM optimizes both memory usage and retrieval speed.}
    \label{fig:idea}
\end{figure}

\section{Introduction}
\label{sec:intro}
Day-to-day experiences often witness lapses in human memory. We may misplace objects within our homes (like searching for a passport), lose track of completed tasks (wondering if salt was added), forget the locations of past activities (where were tickets purchased?), and overlook the status of objects in our surroundings (did I leave the garage door open?). 
Wearable cameras observing human activities from an egocentric perspective have the potential to assist the recall of such episodic memories~\cite{tulving2002episodic} in a wide range of scenarios by continuously keeping track of observed objects and events to alleviate cognitive overload and support the execution of daily tasks~\cite{plizzari2024outlook}.

Towards this direction, the Visual Queries 2D (VQ2D) task has been introduced with the Episodic Memory (EM) benchmark of the Ego4D dataset~\cite{ego4d}. 
VQ2D is formulated as follows: given an image patch of an object query and a long egocentric video, identify a \textit{response track}, the precise spatio-temporal location of the last appearance of the object in the video.
VQ2D has fostered the development of algorithms for episodic memory retrieval in recent years \cite{liu2022reler,feng2024objectnlq,chen2022internvideo,pei2024egovideo,xu2022negativeframesmatteregocentric,xu2023my,ramakrishnan2023spotem,jiang2023single,ramakrishnan2023naq,mai2023egoloc}.
We observe that the current formulation of VQ2D provides for solutions to operate in an \textit{offline} manner.
Specifically, given a visual query at frame $t$, methods search within $\video$ scanning it from current frame $t$ backward up to the moment in which the object is found. 
Arguably, this approach is limited in practical application scenarios in which an always-on wearable device processes several hours of video per day. In fact, offline VQ2D requires the burdensome storage of all past video\footnote{For instance, a 24-hour 8-bit RGB video with 1408x1408 resolution captured at 20FPS \cite{projectaria} can take up to 3.43 TB of uncompressed memory.} and entails significant computational overhead when retrieving multiple queries, requiring the scan of the same video several times. 
We advocate that EM systems should be able to work in an \textit{online} fashion, with limited budgets for computation and storage, reflecting the need to ultimately deploy algorithms on wearable devices. 

We hence propose a new task formulation for EM, Online Visual Query 2D (OVQ2D), in which models are required to work online, processing video frames only once, storing relevant information in a compact memory, and relying on such memory to answer queries whenever they are made by the user. Similarly to offline VQ2D, methods need to return a series of bounding boxes and corresponding past frames where the queried object appears. This enables clear spatio-temporal localization within its visual context, forming an interpretable episode of its last known location.
To assess practical applicability, we compare memory size and computation time, besides mere retrieval performance. 
Note that this online formulation brings several challenges, beyond the need for efficient systems. Specifically: 1) online systems have to identify objects which could be potentially queried in the future without any prior knowledge of their identity, in contrast to offline systems which begin the object search only \textit{after} the query is known;
2) online systems should be able to keep track of the different occurrences of a given object instance and store appropriate response tracks and related frames \textit{in advance}, in contrast to offline systems which initiate a separate search for each query;
3) Online systems should be able to associate each of the potential response tracks with appropriate representations to enable future retrieval, in contrast to offline systems, which can directly compare the query patch with object candidates during the search process.

To tackle OVQ2D, we propose ESOM (Egocentric Stream Object Memory), a principled architecture built on state-of-the-art object detection and tracking, leveraging decades of progress in computer vision.
This formulation adopts an application-driven focus, prioritizing practical deployment and benchmarking over algorithmic novelty.
ESOM processes egocentric video frames online, maintaining a compact, interpretable, and dynamic memory of past objects (see Figure~\ref{fig:idea}). This is achieved with three main components: 1) an \textit{object discovery} module detecting objects of interest in each input frame; 2) an online, long-term \textit{visual object tracker} tracking the discovered objects in the video; 3) a dynamic \textit{object memory} which stores past observations in the form of object tracks with associated visual representations. The object memory is built through an \textit{object memory population} process that coordinates the object discovery and visual tracking modules. It can then be queried by a \textit{query retrieval and localization} process, which efficiently searches the memory and returns the corresponding response track in response to a user query.

Experiments on Ego4D~\cite{ego4d} demonstrate the effectiveness of ESOM over other online methods. Results also show that OVQ2D is challenging, with the best method achieving a success rate of only $\sim$4\%. This modest performance is mainly due to the limitations of current object detection and tracking on real-world egocentric videos. 
When using oracle detection or tracking, ESOM’s success rate rises to 31.91\% and 40.55\%, respectively, and reaches 81.92\% with both.
These results also shed light on the limited performance of current object detection and tracking algorithms when employed in real-world applications, suggesting that more efforts are needed to improve their practical applicability. Unlike standard evaluations that assess these components in isolation, our study offers a competitive benchmark that can be used to assess progress and applicability in real-world downstream tasks.

Overall, the contributions of this work are as follows: 1) we introduce OVQ2D, an online formulation for episodic memory retrieval; 2) we propose ESOM, a novel architecture based on continuous object discovery, tracking, memory update, and query, which is shown to outperform offline methods under competitive memory budgets and brings benefits in terms of runtime; 3) we offer a benchmark based on Ego4D~\cite{ego4d} for OVQ2D, which we also show as a valuable resource to evaluate real-world performance of object detection and tracking algorithms.

\section{Related Work}
\label{sec:relatedwork}

\paragraph{Episodic Memory Visual Query Localization.}
The task of Visual Query Localization was initially formalized in \cite{ego4d}, along with a baseline cascade solution that first searches for the query patch in every frame by means of a Siamese R-CNN \cite{voigtlaender2020siam}, and then tracks it forward and backward in time with a short-term tracker \cite{bhat2020know} to obtain the temporal extent of the response track. 
Based on this formulation, a variety of methodologies explored different key factors such as balancing positive and negative examples~\cite{xu2022negativeframesmatteregocentric}, addressing domain and task biases~\cite{xu2023my}, improving feature extractors~\cite{chen2022internvideo,pei2024egovideo}, and developing end-to-end single-stage frameworks~ \cite{jiang2023single} to improve performance and inference speed. 
Despite the progress, all the aforementioned works are based on the offline assumption. In this paper, we formulate a novel Online Visual Query 2D (OVQ2D) task and offer a first approach and benchmark to tackle the problem.

\paragraph{Object Memory and Cognition.}
Object memory representations have been investigated in robotics~\cite{huang2023out} to support object permanence in manipulation tasks, and in egocentric vision~\cite{plizzari2024spatial} to mimic human spatial cognition in keeping track of object instances when they are out-of-sight.
Recent work~\cite{fan2025videoagent,goletto2024amego}, in particular, explored the use of an object memory to address video understanding tasks such as generic question answering ~\cite{fan2025videoagent} and comprehension of object-based activities in long egocentric videos~\cite{goletto2024amego}. Unlike the aforementioned works, our method is the first to propose and tackle online episodic memory with an object memory, which raises significant challenges due to the need to account for \textit{all} objects appearing in the scene and accurately track object instances in the video. 

\paragraph{Online Processing in Egocentric Vision.}
Previous work tackled the design of algorithms and systems to support online processing, a fundamental requirement for real-world applications on wearable devices~\cite{de2016online,zhao2022testra,miniroad,hoai2014max,sadegh2017encouraging,stergiou2023wisdom,damen2020rescaling,gao2017red,furnari2020rolling,rodin2021predicting,girdhar2021anticipative}, in contrast to a whole line of egocentric vision tasks such as action recognition~\cite{carreira2017quo,feichtenhofer2019slowfast,bertasius2021space}, and temporal action detection~\cite{zhang2022actionformer,liu2024end} which assume that the whole video is available at inference time. 
Episodic memory retrieval tasks, as initially formulated in~\cite{ego4d} are based on the ``offline'' assumption. Limited amount of work considered alternative formulations  resembling an online setting~\cite{barmann2022did}, and a systematic investigation on the topic is still missing.
We address the problem of online episodic memory retrieval targeting the VQL task. Our system achieves online processing abilities by using visual object tracking and a compact queryable memory module to store past experiences.

\section{Online Visual Query 2D Localization}
\label{sec:methodology}
We define Online Visual Query 2D (OVQ2D) as follows.
Given a streaming egocentric video $\video = \big\{ \frame_t \big\}$ of RGB frames $\frame_t$ indexed by time $t$, and a query image patch $\query$ representing the visual appearance of the query object, the goal is to retrieve a visual response track: 
\begin{align}
    \rt = \big\{ (\frame_s,\bbox_s), (\frame_{s+1},\bbox_{s+1}), \dots, (\frame_e,\bbox_e) \big\},
\end{align} 

which is an ordered sequence of (frame, bounding box) pairs $(\frame_j,\bbox_j)$, where $s \leq j \leq e \leq t$. This track determines the spatio-temporal position of the object's last appearance in $\video$, with $s$ and $e$ denoting starting and ending frames, and each box $\bbox_j$ corresponding to frame $\frame_j$. 
This formulation allows for the visualization of $\query$'s spatio-temporal localization within its surrounding context, generating a detailed video snippet that helps users enhance their natural episodic memory.
As in \cite{ego4d}, the goal is to find $\rt$ given $\query$, however, critically, our formulation requires methods to process $\video$ in a streaming fashion, observing frames $\frame_t$ only once. To support future queries, methods can store information in a compact memory.
Figure \ref{fig:idea} provides an illustration of our problem formulation. %

\section{ \frameworkname}
To tackle OVQ2D, we propose the Egocentric Streaming Object Memory (ESOM) architecture. ESOM is based on the object memory $\memory$, a dynamic data structure that memorizes relevant object instances appearing in $\video$. The data structure is populated and queried with two processes: (i) the \textit{object memory population} algorithm that builds $\memory$ online and continuously as $\video$ is being captured; (ii) the \textit{query retrieval and localization} algorithm that retrieves the visual response track $\rt$ related to the  query $\query$ from $\memory$, whenever needed by the camera wearer.  
The continuous interaction between these components allows to dynamically create, update, and query the memory of previously observed objects to support online episodic memory retrieval without the need of storing all the frames of $\video$.%

\subsection{Object Memory}
\label{object_memory}

\paragraph{Representation.}
We define the object memory as a set
\begin{align}
\memory = \big\{ \object_1, \dots, \object_i, \dots, \object_N \big\}
\end{align}
of representations $\object_i$ of the spatio-temporal locations of objects in the frames $\frame_t$ of $\video$. The indexes $1 \leq i \leq N$ denote object instance unique IDs. %
While the representation of $\object_i$ can be arbitrarily abstract, in practice, we set $\object_i$ to be a list $\object_i = \big\{ \obj_{i,t} \big\}$ of records temporally indexed by $t$, with each record having the following form
\begin{align}
\obj_{i,t}  = \big[t, \bbox_{i,t}, \frame_{t}, c_{i,t}\big]
\end{align}
where $\bbox_{i,t} = [b_{i,t}^x, b_{i,t}^y, b_{i,t}^w, b_{i,t}^h]$ denotes the object bounding box, $\frame_{t}$ is the associated RGB frame at time $t$, and $c_{i,t}$ is a \textcolor{black}{binary} label indicating whether the visual content within 
$\bbox_{i,t}$ 
is relevant (i.e., the appearance is distinctive) for the retrieval process. 
If $\obj_{i,t}$ is not visible in $\frame_t$, then $\obj_{i,t} = \emptyset$.
This formulation of the object memory provides an indexed, lightweight, compact, human-readable, and easily queryable representation of objects observed in $\video$ up to the current frame $\frame_t$.
Figure~\ref{fig:memory} visualizes the representation of $\memory$ with respect to the  information of $\video$.

\begin{figure}[t]
    \centering
\includegraphics[width=1\columnwidth]{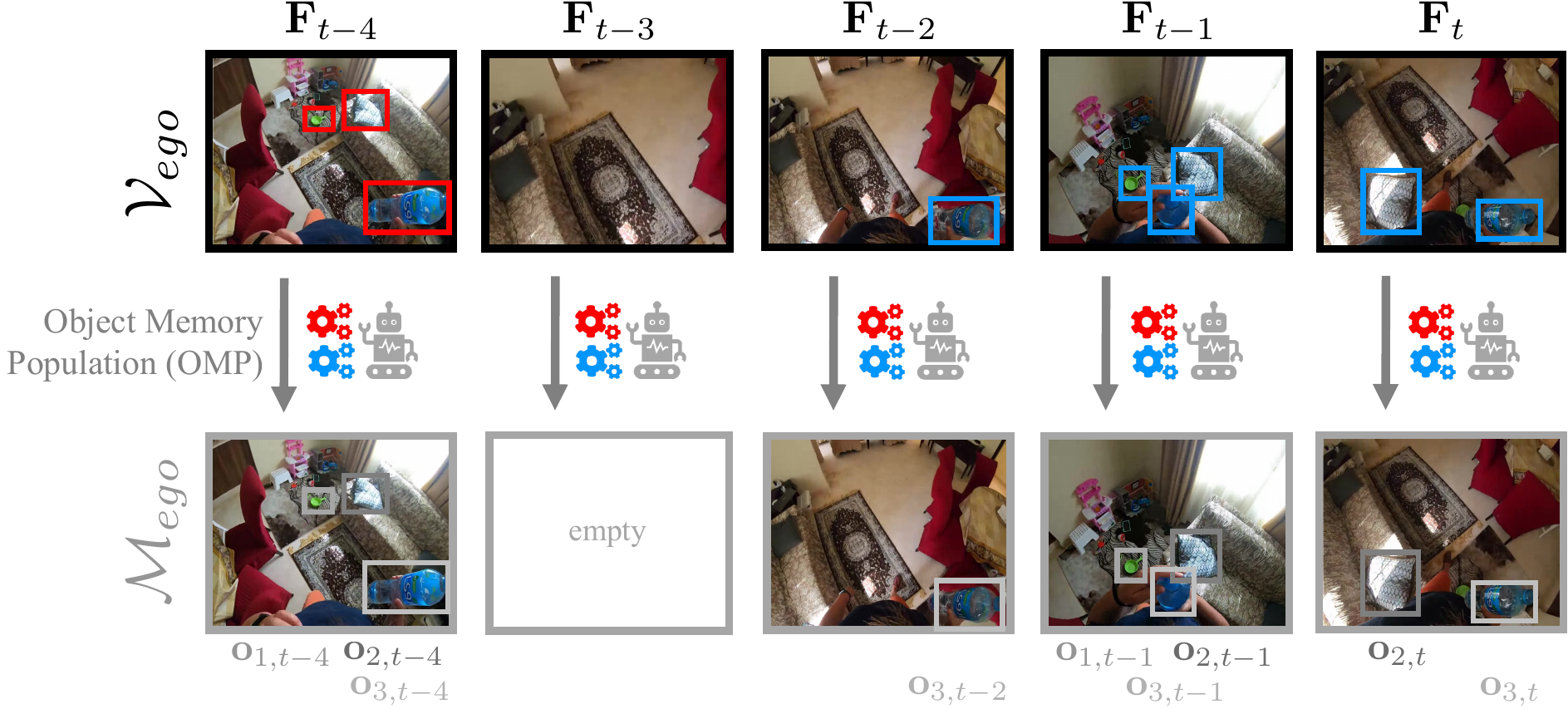}
    \caption{\textbf{From an egocentric video to a memory of objects.} \textcolor{black}{\frameworkacr\ injects the visual information in each frame $\frame_t$ of a video $\video$ into \textcolor[HTML]{7F7F7F}{$\memory$}, an object memory represented as a dynamic list $\object_i$ of tuples $\obj_{i,t}$ composed of instance-based, frame-level bounding boxes $\bbox_{i,t}$, related frames $\frame_t$, and relevance labels $c_{i,t}$.}
    \textcolor[HTML]{7F7F7F}{$\memory$} is built by an \textcolor[HTML]{7F7F7F}{Object Memory Population (OMP)} algorithm which processes (\textcolor[HTML]{FF2140}{\faCogs} \textcolor[HTML]{0F80FE}{\faCogs} \img{images/robot.jpg}) frames $\frame_t$  online.}
    \label{fig:memory}
\end{figure}

\paragraph{Interface.}
Interaction with $\memory$ is achieved through two operations: $.\readop\texttt{()}$, and $.\writeop\texttt{()}$.
The $.\readop\texttt{()}$ operation is used to access information on specific objects stored in $\memory$. Following the object-oriented programming notation, we design .\readop\texttt{()} as an overloaded method that can take as input either an object ID $i$ or a frame ID $t$.  $\memory.\readop\texttt{(} i \texttt{)}$ returns the list of spatio-temporal localizations related to object instance $\object_i$, whereas the $\memory.\readop\texttt{(} t \texttt{)}$ operation returns the spatial information of all objects $\object_i$ that are memorized at time $t$: 
\begin{align}
\big\{ \obj_{j,t} \big\} = \memory.\readop\texttt{(} t \texttt{)}, 1 \leq j \leq N
\end{align}

The $.\writeop\texttt{()}$ operation allows to store new object instances or update existing ones. As before, this operation can be called with different input parameters. The syntax 
\begin{align}
\memory.\writeop\texttt{(}\big[t, \bbox_{k,t}, \frame_{t}, c_{k,t} \big]\texttt{)}
\end{align}
stores with ID $k$ (not yet present in $\memory)$ a new object instance $\object_{k}$ by inserting timestamp $t$, the bounding box $\bbox_{k,t}$, the associated frame $\frame_{t}$, if it is not already in $\memory$, and the label $c_{k,t}$.
The $\writeop$ operation can also be called as 
\begin{align}
\memory.\writeop\texttt{(}\big[t, \bbox_{i,t}, \frame_{t}, c_{i,t} \big]\texttt{)}
\end{align}
to add  $t$, $\bbox_{i,t}$, $\frame_{t}$, and $c_{i,t}$
to object instance $\object_i$ at time $t$ (i.e., we update the representation of $\object_i$). If $\frame_t$ is already stored in $\memory$, \textcolor{black}{no duplicate is stored.}
When $.\writeop\texttt{()}$ updates an object, it checks for a break from the previous timestamp $\hat{t}$ (i.e., $t - \hat{t} > 1$). If occurs, frames $\frame_{\bar{t}}, \bar{t} \leq \hat{t}$ linked to the prior sequence of contiguous boxes are deleted, unless they belong to another memorized object.

\begin{figure}[t]
    \centering
\includegraphics[width=1\columnwidth]{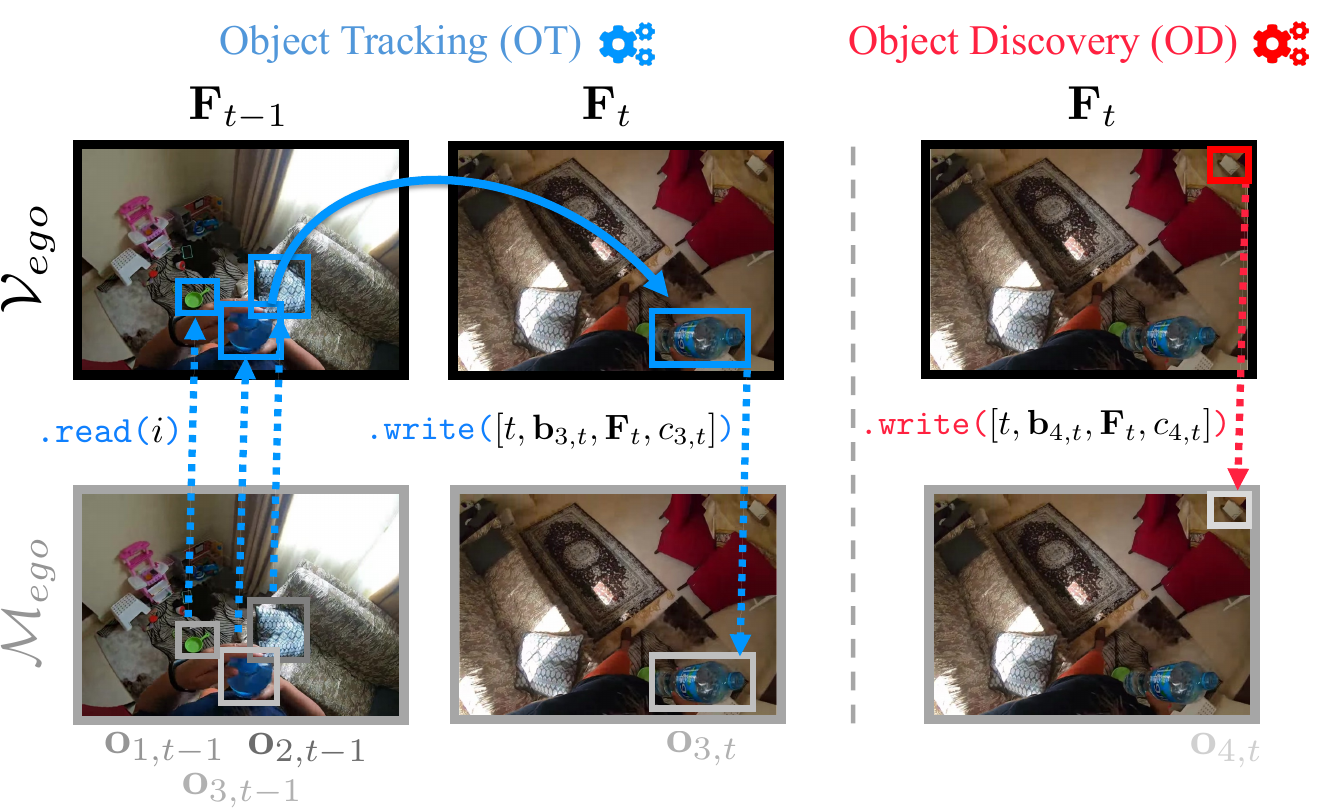}
    \caption{\textbf{Memory population by object tracking and discovery.} 
    \textcolor{black}{The \textcolor[HTML]{0F80FE}{Object Tracking (OT)} (\textcolor[HTML]{0F80FE}{\faCogs}) reads objects $\obj_{i,t-1}$ related to the last frame $\frame_{t-1}$ from \textcolor[HTML]{7F7F7F}{$\memory$} and updates their position in current frame $\frame_t$. 
    In parallel, new objects are detected by the \textcolor[HTML]{FF2140}{Object Discovery (OD)}  (\textcolor[HTML]{FF2140}{\faCogs}) module in frame $\frame_t$.
    Relevance scores $c_{i,t}$ are computed and \textcolor[HTML]{7F7F7F}{$\memory$} is updated.
    }}
    \label{fig:population}
\end{figure}

\subsection{Object Memory Population (OMP)}

The object memory $\memory$ is created and updated by 
a dedicated Object Memory Population (OMP) algorithm designed to process $\video$ online. At each step $t$, 
this algorithm receives as input the latest available frame $\frame_t$ and updates the object memory by executing an Object Tracking (OT) module and an Object Discovery (OD) module. A schematic visualization of the two modules within the OMP algorithm is given in Figure \ref{fig:population}.

\paragraph{Object Tracking Module.} 
The OT module is responsible for updating the localization of objects stored in $\memory$ at previous time steps, maintaining their unique IDs. 
At each frame $\frame_t$, the OT module first retrieves all previously memorized objects $\object_i$ via $\memory.\readop\texttt{(}i\texttt{)}\: \forall i$, then estimates $\bbox_{i,t}$, the position of each object $i$ in frame $t$ by means of a 
visual object tracker \cite{yan2021learning,egotracks}. 
Alongside each $\bbox_{i,t}$, the tracker outputs confidence scores $s_{i,t}$ and only bounding boxes with $s_{i,t} > \lambda_{ot}$ are retained. 
After being filtered by the OT, the remaining bounding boxes and their associated frames are written to memory by calling 
\textcolor{black}{$\memory.\writeop\left(\left[t, \bbox_{i,t}, \frame_{i,t}, c_{i,t} \right]\right)$.}
If no objects are stored in $\memory$, i.e. when $t = 0$ or until the new objects are inserted, the execution of the OT module is skipped. See the supplementary material for more details about the implementation of the OT module.

\paragraph{Object Discovery Module.}
The OD module is responsible for finding objects of interest in frame $\frame_t$ which could be later queried by the user. OD is implemented as an object detector~\cite{wang2405yolov10} which detects a set of $B$ bounding boxes $\bbox_{k,t}$ of objects in the frame, coupled with 
confidence scores $s_{k,t}$, with $1 \leq k \leq B$. After detecting the objects of interest, the OD module calls $\memory.\readop\texttt{(}t\texttt{)}$ to retrieve the bounding boxes of all object instances already present in memory at time $t$. These are objects which may already have been tracked by the OT module at time $t$ from object instances at time $t-1$. The detected boxes are first filtered based on $s_{k,t} > \lambda_{od}$. Those that do not have an IoU $> \lambda_{iou}$ with any of the objects retrieved from the memory, are treated as localizations of new object instances. Filtered $\bbox_{k,t}$ together with the associated frame $\frame_{t}$, are then stored in memory by calling $\memory$.\writeop\texttt{(}$[t, \bbox_{k,t}, \frame_{t}, c_{k,t}]$\texttt{)}, while the others are discarded. See the supplementary material for details.

\paragraph{\textcolor{black}{Retrieval Relevance.}}
To enhance the efficiency of the retrieval stage, each call to the $.\writeop\texttt{()}$ method by OT and OD is preceded by a function that computes relevance scores $c_{i,t}$. These scores identify object patches $\phi_{i,t}$, extracted from $\bbox_{i,t}$ in $\frame_t$, that contain sufficiently distinctive features for accurate instance recognition.
We employ a binary classifier to assess whether each patch $\phi_{i,t}$ meets the quality threshold for effective retrieval (see Supp. material for details). If the classifier determines that a patch is suitable, $c_{i,t} = 1$, otherwise, $c_{i,t} = 0$.

\begin{figure}[t]
    \centering
\includegraphics[width=1\columnwidth]{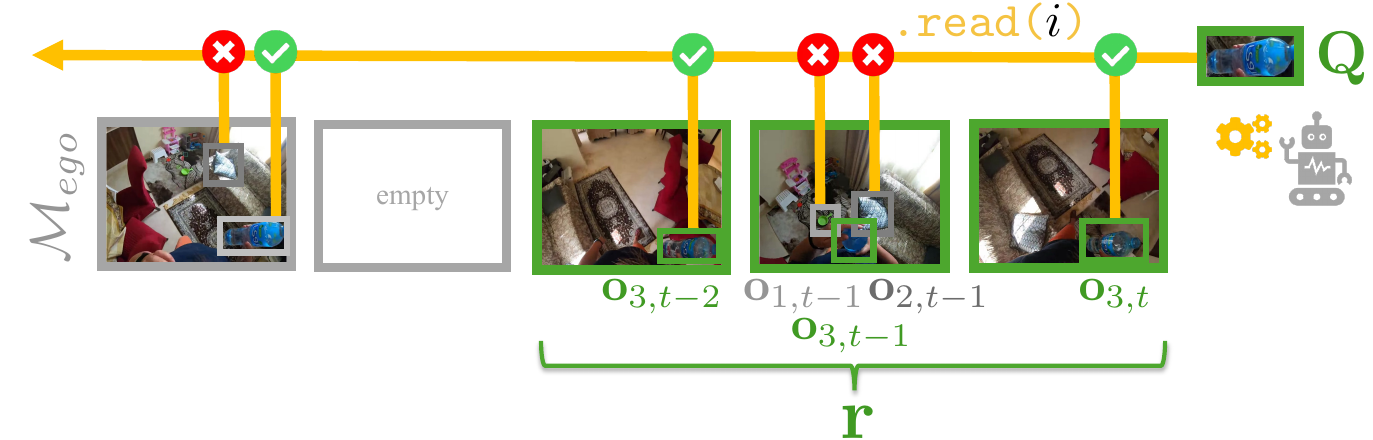}
    \caption{\textbf{Visual query localization by memory retrieval.} 
    When the user provides a visual query \textcolor[HTML]{4DA72E}{$\query$}, the Query Retrieval and Localization algorithm (\textcolor[HTML]{FFC000}{\faCogs} \img{images/robot.jpg}) is triggered. This algorithm compares the representation of \textcolor[HTML]{4DA72E}{$\query$} with the representations of each object $\object_i$ in \textcolor[HTML]{7F7F7F}{$\memory$}. The sequence of contiguous bounding-boxes corresponding to the best-matched (\textcolor[HTML]{72D169}{\faCheckCircle}) object \textcolor{black}{and the associated RGB frames} are retained as the \textcolor[HTML]{4DA72E}{visual response track $\rt$} for \textcolor[HTML]{4DA72E}{$\query$}, while all other objects (\textcolor[HTML]{EA392C}{\faTimesCircle}) are discarded.}
    \label{fig:retrieval}
\end{figure}

\subsection{Query Retrieval and Localization (QRL)}
The Query Retrieval and Localization (QRL) algorithm is responsible for retrieving the visual response track $\rt$ of a visual query $\query$ within the memory $\memory$. 
This algorithm takes $\query$ as input and outputs a sequence of contiguous bounding boxes $\bbox_{i,t}$ and frames $\frame_{t}$ for the object instance $\object_i$ in $\memory$ that visually matches $\query$. The process begins by extracting the query feature representation $\Phi(\query)$ using a feature extraction function $\Phi(\cdot)$. 
Such representation is hence compared against features $\Phi(\phi_{i,t})$ of object states $\obj_{i,t}$ with $c_{i,t}=1$ using a feature similarity function $\Psi(\Phi(\query), \Phi(\phi_{i,t})) \in [0,1]$.
We average similarity scores related to multiple boxes $\obj_{i,t}$ of the same object $i$ into a single score $r_i$.
If the maximum score $r_{\overline{i}}$ across all objects  exceeds the threshold $\lambda_{ret}$, the object \textcolor{black}{$\object_{\overline{i}}$} is considered a match for query $\query$.
We hence return the most recent sequence of contiguous bounding boxes and frames from the spatio-temporal history of \textcolor{black}{$\object_{\overline{i}}$} as the visual response track $\rt$.
Note that this process relies exclusively on $\memory$, and it does not require to access $\video$.
Figure \ref{fig:retrieval} provides an illustration of the process performed by QRL.

\section{Experimental Settings}
\label{sec:experimental}

\begin{table*}[t]
\fontsize{8}{9}\selectfont
	\centering
	
	\tblalternaterowcolors

	\setlength\tabcolsep{.3cm}
	
	\begin{tabular}{ll |l l |c c  c | c c }
		\toprule
		 & Method & OD & OT   & \tAPZS & \stAPZS & \succ & \size & \timesec \\
		\midrule

        1 & AMEGO~\cite{goletto2024amego} & OD$_{hoi-det}$ & OT$_{egostark}$  & 0.0 & 0.0 & 0.67 & 531.6 MB & {1.5} s\\
        
        2 & VideoAgent~\cite{fan2025videoagent}  &OD$_{rt-detr}$ & OT$_{bytetrack}$ &0.02&0.1&3.03& 13.7 GB & 12.1 s\\

        \midrule
        3 & \frameworkacr & OD$_{gdino}$ & OT$_{masa}$   & 0.002 & 0.001  & 0.70 &1.3 GB & 2.5 s \\
        4 & \frameworkacr & OD$_{gdino}$ & OT$_{egostark}$   & 0.04 & 0.002  & 0.72 & 8.8 GB & 24.1 s \\
        5 & \frameworkacr & OD$_{yolo}$ & OT$_{masa}$   & \textbf{0.1} & 0.1  & 3.75 & \textbf{383.6 MB} & \textbf{0.7 s} \\
        6 & \frameworkacr & OD$_{yolo}$ & OT$_{egostark}$   & 0.03 & \textbf{0.2}  & \textbf{4.02} & 1.7 GB & 3.1 s \\

        \midrule
        
         \color{gray}7 & \color{gray}\frameworkacr  & \color{gray}OD$_{oracle}$ & \color{gray}OT$_{egostark}$    & \color{gray}18.3 & \color{gray}5.1 & \color{gray}31.91 & \color{gray}\color{gray}591.4 MB & \color{gray}0.5 s \\
         \color{gray}8 & \color{gray}\frameworkacr  & \color{gray}OD$_{yolo}$ & \color{gray}OT$_{oracle}$    & \color{gray}21.0 & \color{gray}19.4 & \color{gray}40.55 & \color{gray}\color{gray}322.5 MB & \color{gray}0.6 s \\
        \color{gray} 9 & \color{gray}\frameworkacr & \color{gray}OD$_{oracle}$ & \color{gray}OT$_{oracle}$   & \color{gray}73.2 & \color{gray}68.1  & \color{gray}81.92 & \color{gray}\color{gray}507.4 MB  & \color{gray}0.5 s \\

		\bottomrule		
    \end{tabular}

    \caption{\textbf{\textcolor{black}{Comparison of different configurations of \frameworkacr\ for OVQ2D with respect to other online approaches.}} \frameworkacr\   outperforms competitors overall, with OD$_{yolo}$-OT$_{egostark}$ being the best configuration according to success rate and OD$_{yolo}$-OT$_{masa}$ producing much faster inference and more compact memory sacrificing some performance. \textcolor{gray}{Grayed rows (7-9)} report \frameworkacr performance in oracle configurations. (Best results are highlighted in \textbf{bold}.)}
	\label{tab:oncomp}
\end{table*}

\paragraph{Datasets.} We evaluate \frameworkacr on the Ego4D Episodic Memory VQ2D benchmark~\cite{ego4d}, including $433$ hours of video from $54$ scenarios, featuring %
$22K$ 
visual queries spanning $3K$ object categories. 
At present time, this is the only publicly available benchmark for Visual Query Localization. To train components, we also use EgoTracks~\cite{egotracks}, an egocentric video dataset sourced from Ego4D and aligned to the episodic memory benchmark, and EgoObjects~\cite{zhu2023egoobjects}, a large-scale egocentric object detection dataset.

\paragraph{Implementation Details.}
We tested different configurations of the proposed \frameworkacr framework.
In OMP, for the OT component, we used EgoSTARK~\cite{egotracks}, a STARK instance~\cite{yan2021learning} fine-tuned on EgoTracks, as well as the MASA tracker \cite{li2024matching}. 
To evaluate optimal tracking performance, we also considered ground-truth annotations from EgoTracks. We refer to these configurations as OT$_{egostark}$, OT$_{masa}$, OT$_{oracle}$.
For the OD module, we considered the off-the-shelf open-world detector GroundingDino~\cite{liu2023grounding}, an instance of YOLOv10~\cite{wang2405yolov10} trained on curated versions of EgoTracks~\cite{egotracks} and EgoObjects~\cite{zhu2023egoobjects}, and ground truth object detections from EgoTracks. These are denoted as OD$_{gdino}$, OD$_{yolo}$, OD$_{oracle}$.
In QRL, we implemented feature extraction $\Phi(\cdot)$ and similarity $\Psi(\cdot)$  using two configurations: Siam-RCNN with a classification head \cite{voigtlaender2020siam} or DINOv2 with cosine similarity \cite{goletto2024amego,oquab2023dinov2}, referred to as QRL$_{siamrcnn}$ and QRL$_{dino}$.
We obtained best results with OT$_{egostark}$, OD$_{yolo}$ and QRL$_{siamrcnn}$, which identify the default configuration of \frameworkacr\ where not otherwise specified.
See the supplementary material for more implementation details.

\paragraph{Evaluation Protocol and Measures.} 
Each video clip $\video$ in the evaluation dataset is fed to the OMP algorithm to construct the object memory $\memory$. Each visual query $\query$ provided with the clip is then processed using the QRL algorithm, producing a visual response track $\rt$, which we subsequently compare to the annotated ground-truth response track of  bounding-boxes to compute query localization accuracy.
We use the standard VQL evaluation metrics~\cite{ego4d}: 
\tAPZS, measures temporal alignment with the ground-truth at a 0.25 IoU threshold; 
\stAPZS, assesses spatio-temporal precision a 0.25 spatio-temporal IoU threshold; Success (\succ) calculates the proportion of predictions with a minimum 0.05 spatio-temporal IoU with the ground-truth response track.
All metrics are reported in the range [0, 100].
In addition to these standard measures, we evaluate efficiency in terms of storage size (\size) in multiples of bytes (MB, GB) needed to store information that enables query localization, and retrieval time (\timesec) in seconds (s) or minutes (m) that is required to localize a visual query.

\section{Results}

\subsection{Online Visual Query 2D (OVQ2D) Benchmark}

Table \ref{tab:oncomp} benchmarks different configurations of ESOM (3-6) against two baselines (1-2) that apply our QRL method on object memory representations recently proposed for tasks other than Online Episodic Memory.
AMEGO~\cite{goletto2024amego} is designed for interaction-based egocentric video understanding and focuses on objects that the camera wearer has interacted with by hand, whereas VideoAgent~\cite{fan2025videoagent} tackles video question answering and focuses on generic objects. Since AMEGO and VideoAgent were not originally designed for the OVQ2D task, their baselines in Table \ref{tab:oncomp} are adapted by integrating their respective object detection and tracking components—HOI-detector \cite{shan2020understanding} and EgoStark for AMEGO, RT-DETR \cite{zhao2024detrs} and ByteTrack \cite{zhang2022bytetrack} for VideoAgent—with our memory and retrieval modules.
Alongside real-world configurations, in (7-9), we also report baselines using oracular detection (7), object tracking (8) and both (9).
Best results are obtained by the proposed ESOM when state-of-the-art detector (OD$_{yolo}$) and tracker (OT$_{egostark}$) are considered. Besides being competitive in terms of query retrieval accuracy (see columns 4-6), this configuration brings a fast retrieval of 3.1 s and a compact memory of 1.7 GB. Among competitors, VideoAgent achieves a lower success rate (3.03 vs 4.02) with much larger memory \textcolor{black}{13.7 vs 1.7 GB} and slower retrieval time (12.5 vs 3.1 s), while AMEGO, focusing only on hand-interacted objects, achieves a very low success rate of 0.67.
The choice of detector and tracker greatly affects the performance of ESOM. Indeed, using the generic object detector OD$_{gdino}$, rather than our optimized OD$_{yolo}$ leads to very limited success rates, regardless of the tracker. 
OT$_{masa}$ and OT$_{egostark}$ also lead to different results, with OT$_{egostark}$ bringing the best performance, while OT$_{masa}$ allowing to obtain faster retrieval 
and a more compact memory 
sacrificing some retrieval accuracy. 
It is worth noting that the OVQ2D benchmark is very challenging for current approaches, with the best methods achieving a success rate of just 4.02 and very low object localization scores overall.
The oracular results in rows 7-9 show that this is mainly due to the limited effectiveness of current object detection and tracking approaches in real-world egocentric videos.
Indeed, pairing OT$_{egostark}$ with ground truth detections (row 7) brings to a success rate of 31.91, pairing a OD$_{yolo}$ object detection with an oracular tracker leads to a success rate of 40.55, while using both oracular detection and tracking leads to 81.92.
Overall, these results highlight the potential of ESOM to tackle OVQ2D, but also highlight that efforts should be devoted to improve object detection and tracking.
While prior studies highlight challenges in egocentric object detection~\cite{zhu2023egoobjects} and tracking~\cite{dunnhofer2021first,dunnhofer2023visual,egotracks,dunnhofer2025}, algorithms are typically evaluated in isolation~\cite{zhu2023egoobjects,egotracks}, which we argue may misrepresent their real-world reliability. 
Beyond OVQ2D, this benchmark, including its oracular baselines, can serve as a principled testbed to support the development and evaluation of object detection and tracking in real-world, downstream settings, rather than in isolation.

\begin{table}[t]
\fontsize{7}{8}\selectfont
	\centering
	
	\tblalternaterowcolors

	\setlength\tabcolsep{.2cm}
	
	\begin{tabular}{l |c  |c c }
		\toprule
		 Method & \succ & \size & \timesec \\
		\midrule
        \multicolumn{4}{c}{\textbf{Offline Methods (VQ2D)}} \\
        \midrule
         SiamRCNN + KYS \cite{ego4d}  & 39.80 & 12.1 GB &  8.3 m \\

         STARK \cite{yan2021learning}  & 18.70 & 12.1 GB &  45 s  \\

         SiamRCNN \cite{voigtlaender2020siam}  & 43.24 & 12.1 GB & 8.3 m \\
        
         CocoFormer \cite{xu2023my}  & 48.37 & 12.1 GB & 8.3 m  \\

         VQLoC \cite{jiang2023single}  & 55.89 & 12.1 GB & 41 s  \\

         PRVQL \cite{fan2025prvql}  & 57.93 & 12.1 GB & 50 s  \\

        \midrule
        \multicolumn{4}{c}{\textbf{\frameworkacr (OVQ2D)}} \\
        \midrule
         \color{gray}\frameworkacr\ - OD$_{oracle}$ - OT$_{oracle}$  & \color{gray}81.92 & \color{gray}507.4 MB  & 
         \color{gray}0.5 s \\

		\bottomrule		
    \end{tabular}

    \caption{\textbf{Comparison of \frameworkacr\ in OVQ2D versus offline VQ2D methods.} 
    Note that VQ2D is much less challenging than OVQ2D, hence success rate is not directly comparable and  
    these results are reported to discuss the differences between the two approaches.
    \frameworkacr\ offers significant savings in memory storage and enhances the efficiency of query localization.}
	\label{tab:offoncom}
\end{table}

\subsection{Comparison with Offline VQ2D}
Table \ref{tab:offoncom} shows state-of-the-art offline VQ2D methods. It is important to note that offline VQ2D and our online formulation are not directly comparable, OVQ2D is a much harder task in which methods do not know have access to the queried object in advance; hence these results are reported mainly to discuss the differences between the two approaches. Current offline methods require storing the entire video sequence, resulting in significant memory usage\footnote{We consider a fixed memory of 12.1 GB to store a 5-minutes clip as required in the EGO4D benchmark. Longer clips would take more space.}. ESOM reduces memory needs by up \textcolor{black}{7$\times$} (\textcolor{black}{1.7 vs 12.1GB}) and increases inference speed by 13$\times$ when compared to VQLoC (\textcolor{black}{3.1 vs 41 s}). Memory and computational savings are even more significant when considering the oracular configuration, reported for reference, with \textcolor{black}{24$\times$} less memory and \textcolor{black}{80$\times$} increased inference speed. Interestingly, ESOM is competitive even in terms of success rate in oracular settings (81.92 vs 55.89 of VQLoC), which highlights the potential of OVQ2D and ESOM when tracking and detection are solved.

\subsection{Detailed Analysis of ESOM and Ablations}

\begin{table}[t]
\fontsize{6}{6}\selectfont
	\centering

    \setlength\tabcolsep{.03cm}

\begin{tabular}{ l l |l|c c c|c} 

\toprule
\multicolumn{2}{l|}{OMP Configuration} & \multirow{2}{*}{QRL} & \multirow{2}{*}{\tAPZS} & \multirow{2}{*}{\stAPZS} & \multirow{2}{*}{\succ} & \multirow{2}{*}{\timesec} \\
 OD & OT & &  &  &  &  \\

\midrule

 \rowcolor{tblrowcolor2} \cellcolor{white} \multirow{2}{*}{OD$_{oracle}$} & \cellcolor{white} \multirow{2}{*}{OT$_{oracle}$} & QRL$_{siamrcnn}$ & \textbf{73.2}  & \textbf{68.1} & \textbf{81.92} & \textbf{0.5 s} \\ 

&& QRL$_{dino}$  & 64.4  &	57.7 &	76.00 & 0.6 s\\ 

 \midrule
 
\rowcolor{tblrowcolor2} \cellcolor{white} \multirow{2}{*}{OD$_{yolo}$} & \cellcolor{white} \multirow{2}{*}{OT$_{oracle}$} &  QRL$_{siamrcnn}$	& \textbf{21.0}		& \textbf{19.4}	& \textbf{40.55} & \textbf{0.4 s} \\ 
& & QRL$_{dino}$&	20.6 &	12.1&35.53 & 0.4 s\\ 

\midrule

\rowcolor{tblrowcolor2} \cellcolor{white} \multirow{2}{*}{OD$_{oracle}$} & \cellcolor{white} \multirow{2}{*}{OT$_{egostark}$} &  QRL$_{siamrcnn}$ 	&\textbf{18.3}		&\textbf{5.1}	&\textbf{31.91} & \textbf{0.5 s} \\ 

&& QRL$_{dino}$ 	&\textbf{18.3}		&\textbf{5.1}	&30.68 & 0.5 s\\ 

\midrule

\rowcolor{tblrowcolor2} \cellcolor{white} \multirow{2}{*}{OD$_{yolo}$} & \cellcolor{white} \multirow{2}{*}{OT$_{egostark}$} &  QRL$_{siamrcnn}$  &\textbf{0.02}&\textbf{0.2} & \textbf{4.02} & \textbf{3.1 s} \\ 

&& QRL$_{dino}$  &\textbf{0.02} &\textbf{0.2}&3.79 & 3.4 s\\

\bottomrule		
\end{tabular}

\caption{\textbf{Effect of the QRL algorithm on performance.} Performance of QRL methods across various OMP configuations.
QRL$_{siamrcnn}$ consistently achieves the highest accuracy, making it the best-performing model for query retrieval in OVQ2D. 
(Best results are highlighted in \textbf{bold}).
} 
\label{tab:qrl}
\end{table}

\paragraph{Effect of QRL algorithm on performance.} Table \ref{tab:qrl} compares the performance of the QRL algorithm under the two considered implementations.
QRL$_{siamrcnn}$ consistently performs best at matching a visual query with a memory of stored representations. Across all memory constructions produced by various OMP configurations, QRL$_{siamrcnn}$ demonstrates superior query localization scores, emphasizing its strength and adaptability for the OVQ2D task.

\paragraph{Filtering visual representations helps retrieval accuracy and time.} 
We evaluated the impact of computing relevance scores $c_{i,t}$ via the binary classifier before the .\writeop\texttt{()} step on QRL accuracy and speed (Table \ref{tab:classifier}). 
Without relevance filtering, all stored object representations are considered during retrieval, increasing computation time as more similarities must be evaluated. 
Filtering also improves the downstream OVQ2D performance by discarding suboptimal object representations.

\begin{table}[t]
\fontsize{8}{7}\selectfont
	\centering

    \setlength\tabcolsep{.06cm}

\begin{tabular}{ l l c|c c c|c} 

\toprule
\multicolumn{2}{l}{OMP Configuration} & Retr. & \multirow{2}{*}{\tAPZS} & \multirow{2}{*}{\stAPZS} & \multirow{2}{*}{\succ} & \multirow{2}{*}{\timesec} \\
 OD  & OT  & Rel. &  &  &  &  \\
\midrule

 \rowcolor{tblrowcolor2} \cellcolor{white} \multirow{2}{*}{OD$_{oracle}$} & \cellcolor{white} \multirow{2}{*}{OT$_{oracle}$} & \cmark & \textbf{73.2}  & \textbf{68.1} & \textbf{81.92} &\textbf{0.5 s} \\ 

&& \xmark  & 71.5  &	67.4 &	81.37 & 4.6 s \\ 

 \midrule
 
\rowcolor{tblrowcolor2} \cellcolor{white} \multirow{2}{*}{OD$_{yolo}$} & \cellcolor{white} \multirow{2}{*}{OT$_{oracle}$} &  \cmark	&\textbf{21.0}		&\textbf{19.4}	& \textbf{40.55}& \textbf{0.4 s}\\ 
& & \xmark &	18.3 &	18.1& \textbf{40.55}& 5.3 s\\ 

\midrule

\rowcolor{tblrowcolor2} \cellcolor{white} \multirow{2}{*}{OD$_{oracle}$} & \cellcolor{white} \multirow{2}{*}{OT$_{egostark}$} &  \cmark	&\textbf{18.3}		&\textbf{5.1}&\textbf{31.91} & \textbf{0.5 s} \\ 

&& \xmark 	&17.4		&5.0&29.78 &24.2 s \\ 

\midrule

\rowcolor{tblrowcolor2} \cellcolor{white} \multirow{2}{*}{OD$_{yolo}$} & \cellcolor{white} \multirow{2}{*}{OT$_{egostark}$} &  \cmark &0.02& \textbf{0.2} & \textbf{4.02} & \textbf{3.1 s}\\ 

&& \xmark & \textbf{0.03} &0.1&2.68 &24.8 s\\ 
\bottomrule		
\end{tabular}

\caption{\textbf{Ablating the use of the retrieval relevance over different OMP configurations.} 
Using relevant visual representations for retrieval (Retr. Rel.) improves \textcolor{black}{time and accuracy} of QRL.
(Best results in \textbf{bold}). } 

\label{tab:classifier}
\end{table}

\begin{table*}[t]
\fontsize{8}{7}\selectfont
	\centering

    \setlength\tabcolsep{.21cm}

\begin{tabular}{ l c c | l c c | c c c | c c } 

\toprule

OMP Config. & \multirow{2}{*}{\ap} & \multirow{2}{*}{\rec} & OMP Config. & \multirow{2}{*}{\hota} & \multirow{2}{*}{\assa} & \multirow{2}{*}{\tAPZS} & \multirow{2}{*}{\stAPZS} & \multirow{2}{*}{\succ} & \multirow{2}{*}{\size} & \multirow{2}{*}{\timesec}  \\

 OD  &  & & OT  &  &  &  &  & \\
\midrule

 \rowcolor{tblrowcolor2} \cellcolor{white}  & \cellcolor{white}  & \cellcolor{white}  & OT$_{oracle}$ & \textcolor{gray}{100.0} & \textcolor{gray}{100.0} & \textcolor{gray}{73.2}  & \textcolor{gray}{68.1} & \textcolor{gray}{81.92} & \textcolor{gray}{507.4 MB} & \textcolor{gray}{0.5 s} \\ 

  OD$_{oracle}$ & \textcolor{gray}{100.0} & \textcolor{gray}{100.0} & {OT$_{masa}$} & \textbf{50.1} & 25.2 & 2.8 & 1.6	& 9.38 & 1.3  GB & 7.2 s\\ 

  \rowcolor{tblrowcolor2} \cellcolor{white} & \cellcolor{white}  & \cellcolor{white}  & {OT$_{egostark}$} & 21.7 & \textbf{34.7} &\textbf{18.3}		&\textbf{5.1}	&\textbf{31.91} &\textbf{591.4 MB} & \textbf{0.5 s} \\

 \midrule

 & & & OT$_{oracle}$ & \textcolor{gray}{84.8} & \textcolor{gray}{94.3} & \textcolor{gray}{21.0}  & \textcolor{gray}{19.4} & \textcolor{gray}{40.55} & \textcolor{gray}{322.5 MB} & \textcolor{gray}{0.4 s} \\ 

  \rowcolor{tblrowcolor2} \cellcolor{white} OD$_{yolo}$ & \cellcolor{white} 13.6 &  \cellcolor{white} 30.7 & {OT$_{masa}$} & \textbf{6.6} & 11.6  & \textbf{0.1} & \textbf{0.1}	& 3.75 & \textbf{383.6 MB} & \textbf{0.7 s} \\ 
  \rowcolor{tblrowcolor2} \cellcolor{white} & \cellcolor{white} & \cellcolor{white} & {OT$_{egostark}$} & 6.1 & \textbf{15.1} & 0.03 & 0.2 & \textbf{4.02} & 1.7 GB & 3.1 s \\

\midrule

 \rowcolor{tblrowcolor2} \cellcolor{white} & \cellcolor{white} & \cellcolor{white} &  OT$_{oracle}$ & \textcolor{gray}{ 87.5}& \textcolor{gray}{94.3}& \textcolor{gray}{15.7} & \textcolor{gray}{12.2} & \textcolor{gray}{33.33}& \textcolor{gray}{355.1 MB} & \textcolor{gray}{0.5 s} \\ 

   OD$_{gdino}$ & 4.5 & 25.7 & {OT$_{masa}$} & 3.6 & 10.5  & 0.0 & 0.0	& 0.70 & \textbf{1.3 GB} & \textbf{2.5 s} \\ 

  \rowcolor{tblrowcolor2} \cellcolor{white} & \cellcolor{white} & \cellcolor{white} & {OT$_{egostark}$} & \textbf{5.7} & \textbf{15.1} & \textbf{0.04} & 0.0 & \textbf{0.72} & 8.7 GB & 24.1 s \\

\bottomrule		
\end{tabular}

\caption{\textbf{Effects of OD and OT modules on performance.} Different OD and OT model implementations determine different behaviors of the OMP algorithm, resulting in different memory representations. This, in turn leads to different QRL results. Performance of methods with oracular object tracking is reported \textcolor{gray}{in gray}. (Best results are in \textbf{bold}).}
\label{tab:discoverytracking}
\end{table*}

\paragraph{Effect of OD and OT modules.} 
Table \ref{tab:discoverytracking} reports ESOM’s performance across different OD and OT configurations, including module-specific and VQL task metrics: \ap\ and \rec\ for object detection\footnote{All detections are merged into a single ``object'' class.}, \hota\ and \assa\ for object tracking \cite{luiten2021hota}, and \%(s)tAP$_{25}$, Succ., Size, and Time for downstream OVQ2D.

The OD module has a noticeable impact on performance, 
as weaker detectors might miss important queries and add objects into memory that are unlikely to be queried later. 
OD$_{gdino}$, designed for open-world detection, has low recall and AP for annotated queries, leading to weak retrieval even with OT$_{oracle}$. In contrast, OD$_{yolo}$, trained to detect queried objects, enables better memorization and improves QRL accuracy.
Changes in the detector impact memory size and processing time; OD$_{gdino}$ detects more objects, expanding memory and increasing query localization time.

OT modules have a more significant impact on performance. replacing the oracular tracker OT$_{oracle}$ with OT$_{egostark}$ while keeping OD$_{oracle}$  brings Succ. down from 81.92 to 31.91. We observe an inverse correlation between downstream and general tracking performance. For instance, OT$_{masa}$ achieves the highest HOTA score (50.1 vs. 21.7), yet yields significantly lower Succ. (9.38 vs. 31.91). This stems from HOTA being the geometric mean of DetA (detection accuracy within individual frames) and AssA (association accuracy across time) \cite{luiten2021hota}. In our evaluations, DetA tends to be consistently lower than AssA — for example, OD$_{oracle}$–OT$_{egostark}$ shows 13.7 vs. 34.7, and OD$_{yolo}$–OT$_{egostark}$ yields 2.5 vs. 15.1—ultimately reducing overall HOTA scores. OVQ2D requires higher AssA than DetA, as accurate object association is essential for constructing coherent and temporally consistent response tracks in $\memory$. Incorrect associations lead to fragmented or duplicated tracklets, which DetA penalizes. OD$_{oracle}$–OT$_{masa}$ stands out as an exception, relying on ground truth detections and thus isolating the role of association, which explains its strong HOTA despite limited downstream success. 
Being explicitly optimized for egovision, OT$_{egostark}$ leads
to the best downstream performance.

\paragraph{Behavior of ESOM with different amounts of memory.}
Figure \ref{fig:accuracy_mem_size_var} shows retrieval performance, storage space, and retrieval time as 25\%, 50\%, 75\%, and 100\% of videos are observed with different OMP configurations. QRL performance declines slightly with video length and remains steady with oracular OD and OT, while minor drops occur with OD$_{oracle}$ and OD$_{egostark}$, underscoring ESOM’s scalability. Storage and retrieval time generally increase over time; however, memory size remains stable with OD$_{oracle}$ since most objects appear within the first 25\% of the video. OD$_{yolo}$ - OT$_{egostark}$ significantly impacts storage and retrieval due to duplicate objects from weaker tracking.

\paragraph{Qualitative Results.} Figure \ref{fig:qualex} shows a qualitative example of QRL retrieving and localizing a visual query within two memory representations.

\begin{figure}[t]
    \centering
\includegraphics[width=\columnwidth]{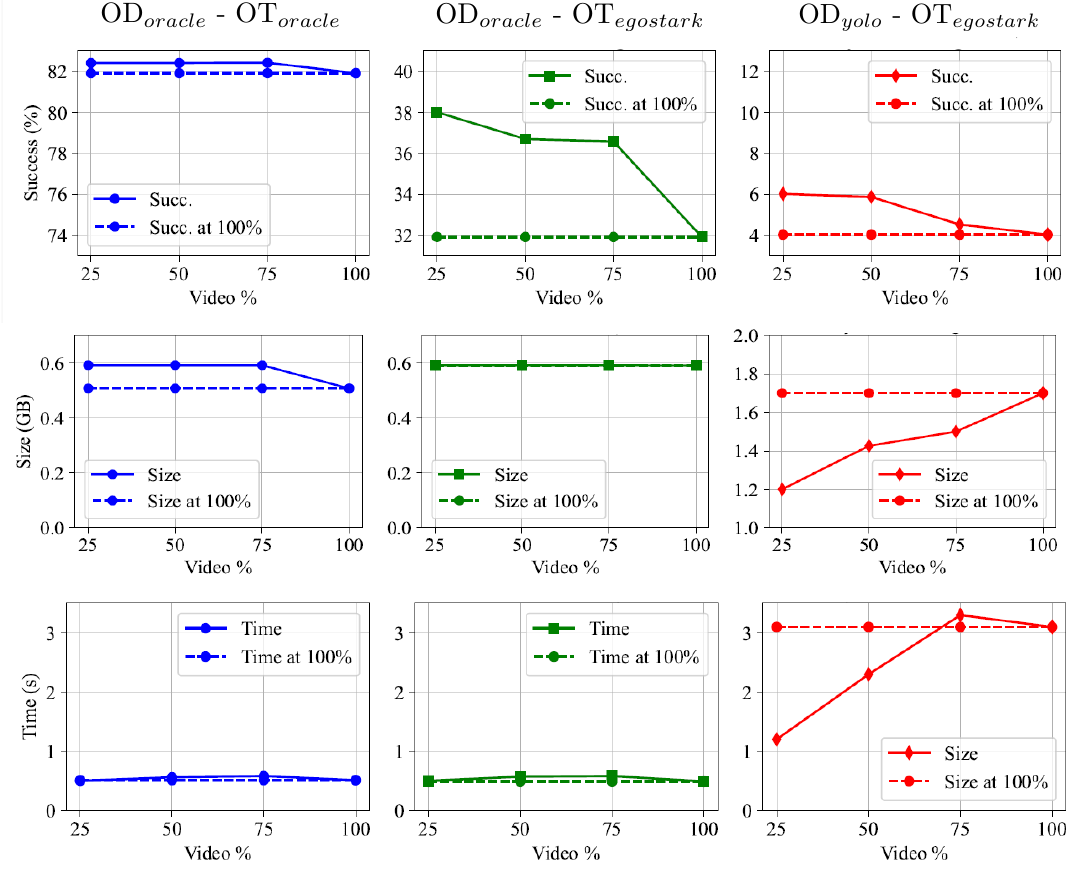}
    \caption{\textbf{ESOM scales well while increasing memory size.} Plots show how success score, storage space, and retrieval time change when building memory from progressively processed video segments. Results for three different OMP configurations.}
    \label{fig:accuracy_mem_size_var}
\end{figure}

\begin{figure}[t]
    \centering
 \includegraphics[width=\columnwidth]%
 {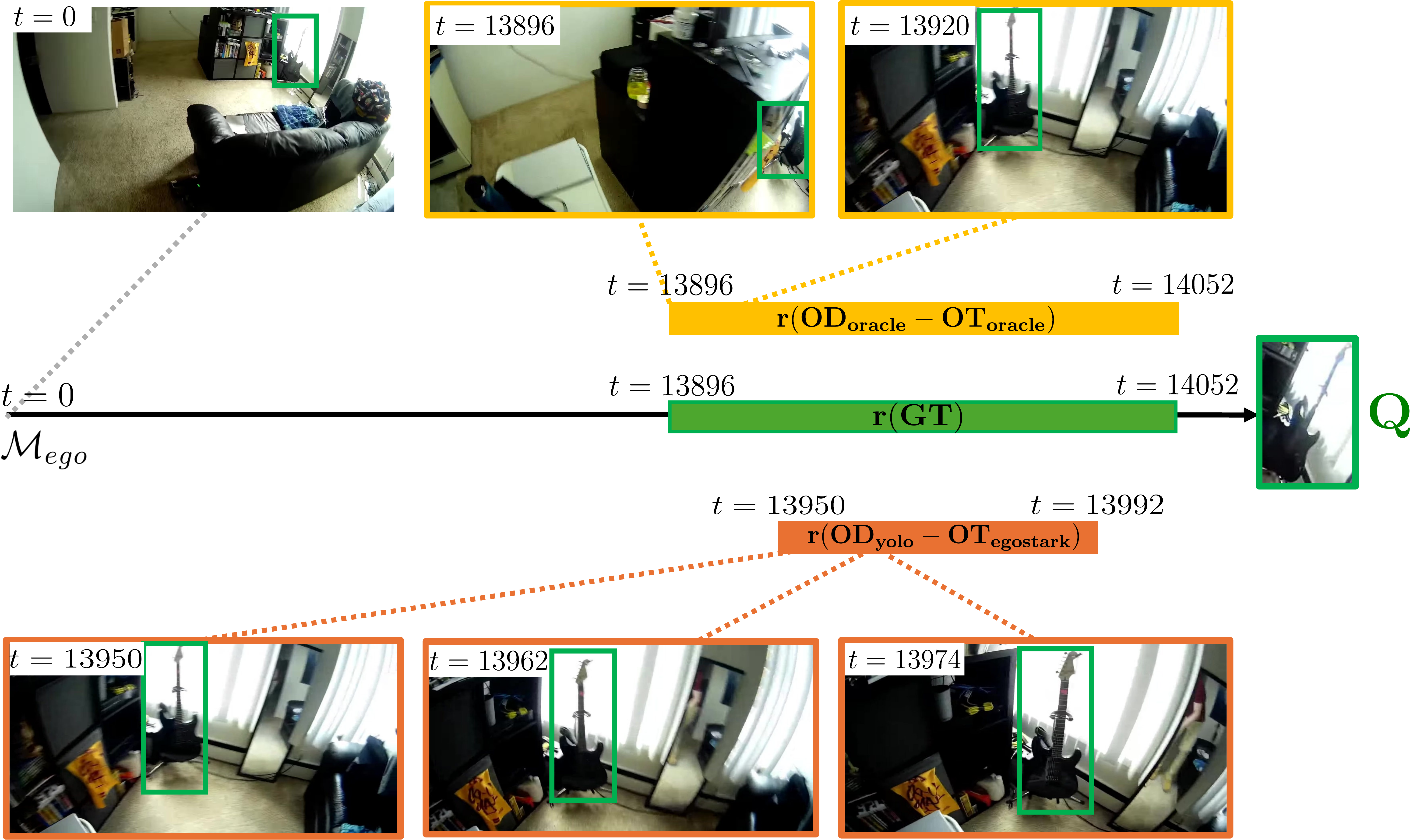}
 \caption{\textbf{Qualitative example.}  Successful retrieval of query  \textcolor[HTML]{4DA72E}{$\query$} from {$\memory$} using the oracle OMP algorithm (\textcolor[HTML]{CC9900}{OD$_{oracle}$ - OT$_{oracle}$}) and the real-world OMP (\textcolor[HTML]{ED7D31}{OD$_{yolo}$ - OT$_{egostark}$}). 
Sample frames from each track are shown with corresponding \textcolor[HTML]{4DA72E}{bounding boxes} whose patches $\phi_{i,t}$ were matched during retrieval. The frame $t=0$ corresponds to the moment of the object's discovery. The \textcolor[HTML]{4DA72E}{green bar} visualizes the temporal extent of the ground-truth track, with timestamps indicating its duration.
 }
    \label{fig:qualex}
\end{figure}

\section{Conclusions}

We introduced the novel task of Online Visual Queries 2D (OVQ2D), where a video is processed in a streaming fashion and an object query must be localized without access to the original video. To address it, we proposed ESOM (Egocentric Stream Object Memory), an online approach that detects, tracks, and stores essential objects to support efficient visual query retrieval and localization. ESOM significantly optimizes both memory and computational efficiency, reducing memory usage by up to \textcolor{black}{7$\times$}, shortening retrieval times from minutes to a few seconds. %
On the Ego4D dataset ESOM outperforms other online methods and shows further gains when object detection and tracking components are improved. 
Besides establishing
a benchmark for OVQ2D, we also contribute a principled
evaluation, based on a real downstream task, for object detection
and tracking algorithms for egovision applications.

\paragraph{Acknowledgments.} Progetto PRIN 2022 PNRR - ``Tracking in Egovision for Applied Memory (TEAM)'' Codice P20225MSER\_001. Codici CUP G53D23006680001 (Università di Udine) e E53D23016240001 (Università di Catania). This research has been funded by the European Union, NextGenerationEU – PNRR M4 C2 I1.1 RS Micheloni.
MD received funding from the European Union’s Horizon Europe research and innovation programme under the Marie Skłodowska-Curie grant agreement n. 101151834 PRINNEVOT (CUP G23C24000910006).

{
    \small
    \bibliographystyle{ieeenat_fullname}
    \bibliography{main}
}

\clearpage
\setcounter{page}{1}
\maketitlesupplementary
\appendix

\noindent
This document reports implementation details on the proposed \frameworkacr\ methodology, along with additional results and discussion. Specifically, Section~\ref{sec:esom} explains the ESOM architecture by providing details of the Object Memory representation (sub-section~\ref{memory_config}), the Object Tracking (OT) module (sub-section~\ref{sec:supp_OT}), on the Object Discovery (OD) module (Section~\ref{sec:supp_OD}), and on the Query Retrieval and Localization (QRL) algorithm (sub-section~\ref{sec:supp_qrl}). Section ~\ref{sec:baselines} reports details on the implementation of the baselines methods on AMEGO~\cite{goletto2024amego} and VideoAgent \cite{fan2025videoagent}. Section~\ref{sec:supp_dataset} reports information on the evaluation dataset and Section~\ref{sec:supp_qualitative} reports additional and qualitative results.

\section{Implementation Details of ESOM}
\label{sec:esom}

\subsection{Object Memory}
\label{memory_config}

\paragraph{Visual representation filter.}
    \textcolor{black}{To improve retrieval time and accuracy}, the .\writeop\texttt{()} operation incorporates a filtering mechanism to determine \textcolor{black}{whether the visual content within $\bbox_{i,t}$, $\phi_{i,t}$, is relevant for the retrieval process}. This filtering process relies on a binary classification function that assigns a label $c_{i,t}$ to each visual representation $\phi_{i,t}$, assessing its quality for effective query retrieval.\\
  \textcolor{black}{The design of $\memory$, which includes information about the quality of data for retrieval, was guided by experimental results. Specifically, we tested various filtering approaches to determine their effectiveness.}
\begin{itemize}
    \item MR1* (\emph{design as reported in Section 4.1 of the main paper}):  \textcolor{black}{ for each object $\object_i$, the filter assigns $c_{i,t} = 1$  at the moment of object discovery, as well as to two bounding-boxes $\bbox_{i,t}$ and their associated frames $\frame_t$ from the most recent response track. These bounding-box and frame pairs are selected as the first and last valid patches identified by the classifier within the latest sequence of bounding boxes.  For all other frames in temporal locations not mentioned, the filter set $c_{i,t} = 0$.}
    
    \item MR2: \textcolor{black}{for each object $\object_i$,  the filter assigns $c_{i,t} = 1$  at the moment of object discovery, as well as to all the bounding-boxes $\bbox_{i,t}$ and their associated frames $\frame_t$ from the most recent response track, considered useful by the classifier. This implementation was conducted to assess the impact of exploiting more object visual information for QRL.}
   
    \item MR3: \textcolor{black} {for each object $\object_i$,  the filter assigns $c_{i,t} = 1$  at the moment of object discovery. For all other frames in temporal locations not mentioned, the filter set $c_{i,t} = 0$. This implementation was conducted to assess the impact of exploiting less object visual information for QRL, thus reducing the retrieval time of $\memory$.}
    
    \item MR4: \textcolor{black}{for each object $\object_i$, the filter assigns $c_{i,t} = 1$  to two bounding-boxes $\bbox_{i,t}$ and their associated frames $\frame_t$ from the most recent response track. These bounding-box and frame pairs are selected as the first and last valid patches identified by the classifier within the latest sequence of bounding boxes.  For all other frames in temporal locations not mentioned, the filter set $c_{i,t} = 0$.  This implementation was conducted to assess the impact of considering the object at the moment of discovery in retrieval performances.}
        
\end{itemize}

\textcolor{black}{The OVQ2D performance of these memory configurations built by the Object Memory Population (OMP) algorithm and processed by the Query Retrieval and Localization (QRL) algorithm are reported in Table \ref{tab:memoryrep}. As can be noticed, the proposed memory representation (MR1*) offers the best balance between QRL accuracy and retrieval time.}

\textbf{Classifier implementation details}
    The classification model was implemented using a deep neural network consisting of a ResNet18 \cite{he2016deep} feature extractor and a linear layer, which served as the representation filter. This model was fine-tuned on a dataset of 17,071 visual representations extracted from 250 uniformly sampled clips from the training set of the Ego4D Episodic Memory VQ2D-EgoTracks benchmark \cite{ego4d,egotracks}. 
Specifically, the training set included 8,534 visual representations labeled as 0 (i.e., not suitable for QRL) and 5,153 representations labeled as 1 (i.e., suitable for QRL). The validation set consisted of 2,126 visual representations labeled as 0 and 1,258 representations labeled as 1.

The training and validation representations were labeled as 0 or 1 based on whether they were incorrectly or correctly matched by the QRL algorithm.
For each frame in the 250 selected clips, the representation $\phi_i$ of the annotated visual query $\query_i$ was extracted, and its feature representation $\Phi(\phi_i)$ was computed using \textcolor{black}{Siam-RCNN \cite{voigtlaender2020siam} as $\Phi(\cdot)$. This representation was then compared to $\Phi(\phi_{i,t})$, the feature representation of the patches withing the ground-truth bounding boxes of the tracklet associated with the query $\query_i$, using a classification head \cite{voigtlaender2020siam}}. This comparison produced a similarity score $s_t$ ranging from 0 to 1. Patches were assigned negative labels if $s_t \leq 0.3$ and positive labels if $s_t \geq 0.7$. After labeling, the data was randomly split into training and validation sets, with 80\% allocated to the training set and 20\% to the validation set.

The visual representation classifier was fine-tuned for 50 epochs using Stochastic Gradient Descent (SGD) with a learning rate of 0.01. At convergence, it achieved a test accuracy of ~97\%. The input patches were resized to $224 \times 224$ pixels before being fed into the model.

\begin{table}[t]
\fontsize{7}{7}\selectfont
	\centering

    \setlength\tabcolsep{.03cm}

\begin{tabular}{ l l |l|c c c|c} 

\toprule
\multicolumn{2}{l|}{OMP Configuration} & \multirow{1}{*}{Memory} & \multirow{2}{*}{\tAPZS} & \multirow{2}{*}{\stAPZS} & \multirow{2}{*}{\succ} & \multirow{2}{*}{\timesec} \\
 OD & OT &Representation &  &  &  &  \\

\midrule

 \rowcolor{tblrowcolor2} \cellcolor{white} & \cellcolor{white} & MR1* & 73.2  & 68.1 & \textbf{81.92} & 0.51 s\\ 

&& MR2  &\textbf{73.3}  & \textbf{68.3} & \textbf{81.92}  & 4.27 s\\ 
\rowcolor{tblrowcolor2} \cellcolor{white} \raisebox{1.6ex}[1.6ex]{} & \cellcolor{white}& MR3  & 63.7  &55.1 & 74.33 & \textbf{0.05 s}\\ 
\multirow{-4}{*}{OD$_{oracle}$} &  \multirow{-4}{*}{OT$_{oracle}$} & MR4  & 70.4  &66.2 & 80.35 & 0.32 s \\ 

 \midrule
 
\rowcolor{tblrowcolor2} \cellcolor{white} &  \cellcolor{white}  &  MR1*	& \textbf{21.0}& 19.4	& \textbf{40.55} &0.36 s\\ 

& & MR2& \textbf{21.0}	& \textbf{19.5	}& \textbf{40.55}& 2.71 s\\ 
\rowcolor{tblrowcolor2} \cellcolor{white}& \cellcolor{white} & MR3&20.3 &17.0 & 38.02 & \textbf{0.05 s}\\ 
\multirow{-4}{*}{OD$_{yolo}$} & \cellcolor{white} \multirow{-4}{*}{OT$_{oracle}$} & MR4&20.7& 17.2& 39.05 & 0.21 s\\ 

\midrule

\rowcolor{tblrowcolor2} \cellcolor{white} &  \cellcolor{white}  & MR1* 	&\textbf{0.18}	&\textbf{0.05} &\textbf{31.91} & 0.49 s\\

&&MR2&\textbf{0.18}	&\textbf{0.05} &\textbf{31.91} & 4.98 s\\ 
\rowcolor{tblrowcolor2} \cellcolor{white}& \cellcolor{white} &MR3&	0.16&0.03 & 26.56& \textbf{0.05 s}\\ 
\cellcolor{white} \multirow{-4}{*}{OD$_{oracle}$} & \cellcolor{white} \multirow{-4}{*}{OT$_{egostark}$} &MR4&0.16& 0.04& 29.02&0.29 s\\ 

\midrule

\rowcolor{tblrowcolor2} \cellcolor{white} &  \cellcolor{white}  &  MR1* &\textbf{0.03}&\textbf{0.02} & \textbf{4.02} & 3.08 s\\

&& MR2 &\textbf{0.03}&\textbf{0.02} & \textbf{4.02} & 14.90 s \\ 
\rowcolor{tblrowcolor2} \cellcolor{white}& \cellcolor{white}& MR3 &0.02& 0.03& 1.78 & \textbf{0.05 s} \\
\multirow{-4}{*}{OD$_{yolo}$} & \cellcolor{white} \multirow{-4}{*}{OT$_{egostark}$} & MR4 &0.02&0.05 &2.90 & 0.69 s\\

\bottomrule		
\end{tabular}

\caption{\textbf{Effect of different filtering strategies.} \textcolor{black}{This table illustrates the impact of using different filtering approaches for $\memory$ population on QRL performance and retrieval time. MR1* refers to the representation described in the main paper. As can be noticed, MR1* constitutes the best approach at the intersection of QRL accuracy and retrieval efficiency.}
(Best results are highlighted in \textbf{bold}).
} 
\label{tab:memoryrep}
\end{table}

 \begin{table}[t]
    \centering
    \tblalternaterowcolors
    
    \resizebox{\linewidth}{!}{
    \begin{tabular}{l | l | cccc}
    \toprule
    Tracker & Detector &  \hota\ & \owta\ & \assa\ &  DetA ${\uparrow}$ \\
    \midrule
    
    MASA  & GroundingDino  & 0.036 & \textbf{0.211} & 0.105 & 0.013\\
    MASA  & YOLOv10*  & \textbf{0.066} & 0.080 & 0.116 & \textbf{0.038} \\
    MASA & \textcolor{gray}{Ground-truth} & \textcolor{gray}{0.501} & \textcolor{gray}{0.501} & \textcolor{gray}{0.252} & \textcolor{gray}{1.000} \\
    \midrule
    EgoSTARK  & Hand-Object Detector & 0.034 & 0.121 & 0.102 &  0.012 \\
    EgoSTARK  & GroundingDino  & 0.057 & 0.121  & 0.115 &  \textbf{0.029} \\
    EgoSTARK & YOLOv10*  & 0.061 & 0.188 &  \textbf{0.151} &  0.025 \\

    EgoSTARK & \textcolor{gray}{Ground-truth}    & \textcolor{gray}{0.217} & \textcolor{gray}{0.390} & \textcolor{gray}{0.347} & \textcolor{gray}{0.137} \\
    \midrule
    BoT-SORT & YOLOWorld & 0.014&0.073&0.042&0.005\\
    \midrule
    \textcolor{gray}{Ground-truth} & YOLOv10* & \textcolor{gray}{0.848} & \textcolor{gray}{0.848} & \textcolor{gray}{0.943} &  \textcolor{gray}{0.762}\\
    \textcolor{gray}{Ground-truth} & GroundingDino  & \textcolor{gray}{0.875} & \textcolor{gray}{0.875} & \textcolor{gray}{0.943} &  \textcolor{gray}{0.812} \\

    \bottomrule
    \end{tabular}
    }
    \caption{\textbf{Evaluation of different tracking algorithms for the implementation of the Object Tracking (OT) module.} We evaluated the multi-object tracking (MOT) performance of our implmentations on EgoTracks's validation set. Models marked with "*" are finetuned on EgoTracks \cite{egotracks}. Ground-truth refers to the EgoTracks annotations used to simulate optimal object detection or tracking, and the respective results are reported in \textcolor{gray}{gray}. We found the AssA results to be best indicators for QRL accuracy. (Best results, across real-world tracker-detector combinations, are highlighted in \textbf{bold}.) %
    }
    \label{tab:tracking_eval}
\end{table}

\subsection{Object Memory Population (OMP)}
\label{sec:supp_OT2}
\subsubsection{Object Tracking (OT) module.}

In this section, we provide details regarding the implementation of OT module. In Table \ref{tab:tracking_eval}, we report the performance of the different implementations as measured by standard multiple object tracking (MOT) metrics \cite{luiten2021hota}.

\paragraph{OT$_{egostark}$.}
OT$_{egostark}$ implements a multiple-object tracking approach using multiple instances of single-object trackers. This approach is motivated by the lack of effective multiple-object tracking solutions for egocentric vision, whereas single-object tracking has been extensively studied \cite{dunnhofer2021first,dunnhofer2023visual,egotracks}.
For single-object tracking, we use the STARK framework \cite{yan2021learning}, fine-tuned on the EgoTracks dataset as described in \cite{egotracks}, resulting in the EgoSTARK tracker. To track multiple objects, we instantiate multiple EgoSTARK trackers simultaneously.
A new tracker instance is initialized in frame $\frame_t$ whenever the OD module identifies a valid object, which is then stored in memory. Each bounding box predicted by the OD module serves as a template to initialize a corresponding tracker. Once initialized, a tracker runs across subsequent frames to predict the bounding boxes of the object it was assigned to, thereby updating the spatio-temporal representation of the respective object. 
Due to the challenges of object tracking in egocentric vision \cite{dunnhofer2021first,dunnhofer2023visual,egotracks}, we found different tracker instances to often end in tracking the same object.
In order to prevent duplicate tracker instances, we calculate the IoU between all tracked objects at frame $t$. If two tracking predictions overlap with an IoU greater than $0.5$, the trackers are considered to be duplicate and only the tracker instance and their corresponding prediction with the lower object instance ID (i.e. the older object) is retained.

Furthermore, to prevent incorrect tracking predictions from causing a tracker to deviate from a searching area that contains the target object and wander across the frame, the box used to compute the search region for the subsequent frame is not updated if the confidence score of the EgoSTARK instance falls below 0.4.
This strategy is based on the observation that, in egocentric vision, objects often leave the frame during rapid head movements but typically reappear in a similar location when the head returns to its original position. 
Before writing the OT module predictions to $\memory$, the bounding boxes $\bbox_{i,t}$ with confidence score $s_{i,t} > \lambda_{ot} = 0.5$ are retained.

\paragraph{OT$_{masa}$.}
As an alternative approach, we implemented the state-of-the-art association method MASA \cite{li2024matching} as a multi-object tracker, utilizing object detections provided by the OD module.
We set a score threshold of $0.1$ and limited object comparisons to the last ten frames using the MASA adapter. To evaluate the impact of considering more frames, we tested with 150 previous frames, but this led to a drop in the OWTA metric from $0.2111$ to $0.1496$. This result highlights the challenges the MASA adapter faces with long-term tracking.

\noindent
\textcolor{black}{\paragraph{OT$_{yoloworld}$.} Finally, we investigated the tracking algorithm BoT-SORT \cite{aharon2022bot} included in the  YOLOWorld \cite{cheng2024yolo} implementation.}

\noindent
\paragraph{OT$_{oracle}$} We implement the oracular object tracker as follows: whenever a detected object intersects a GT bounding box from EgoTracks, we put the rest of the track (from that box onward) in memory. EgoTracks extends the Episodic Memory VQ2D benchmark introducing an object track for each query, hence GT tracks from EgoTracks are valid GT VQ2D tracks. When we combine OT$_{oracle}$ with OD$_{oracle}$, we simply insert in memory all object tracks from EgoTracks.

\subsubsection{Object Discovery (OD) module}
\label{sec:supp_OD}
\paragraph{Clustering-based object taxonomy on EgoTracks.}
As detecting objects from egocentric videos is not straightforward~\cite{zhu2023egoobjects}, we tested different object detectors including the off-the-shelf detectors GroundingDino~\cite{liu2023grounding} and the hand-object detector of~\cite{shan2020understanding} (here referred to as Hand-Object Detector) which do not require training on the considered settings, as well as Faster R-CNN R101-FPN~\cite{ren2016faster} and YOLOv10~\cite{wang2405yolov10} instances trained on different data sources, including EgoObjects~\cite{zhu2023egoobjects} and curated versions of EgoTracks~\cite{egotracks}.
Targeting visual object tracking, EgoTracks contains spatially sparse bounding box annotations across contiguous frames (i.e., only few object instances are annotated in each frame). Each object instance is associated to a textual description rather than an object class (e.g., ``a blue bottle'').
To enable the training of object detectors, we first clustered object descriptions with K-Means to obtain an initial set of $1,169$ classes, followed by manual refinement in order to adjust them, leading to a diverse, granular taxonomy comprising $295$ classes. 

\begin{figure}[t]
    \centering
    \includegraphics[width=\columnwidth]{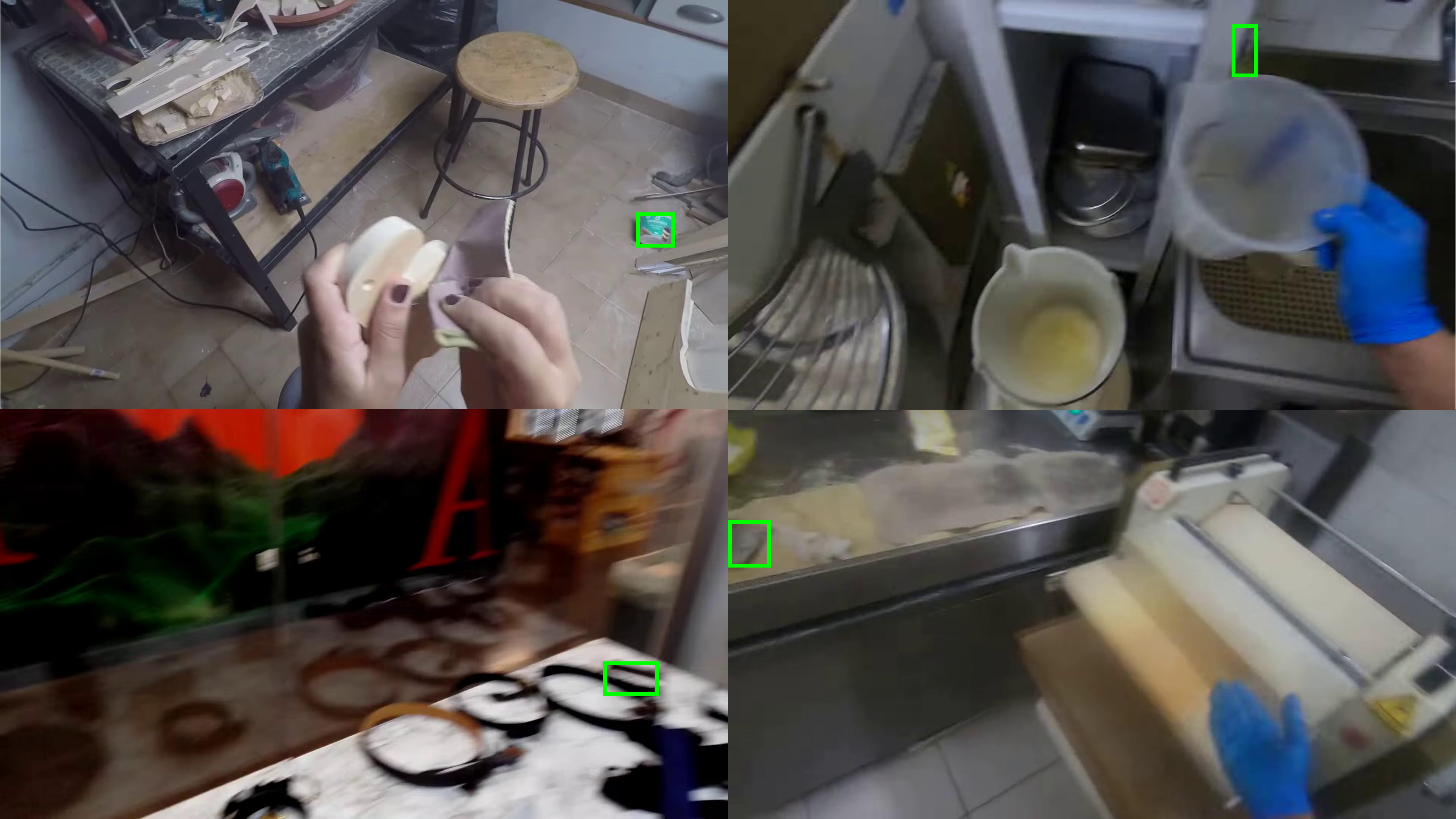}
   
    \caption{\textbf{Visualization of discarded bounding-boxes.} Small-area bounding boxes similar to those shown in this figure have been discarded for the training and evaluation of the module. This is because we found that they lacked sufficient visual information, leading to an increased number of false positive detections. These detections provided no useful data for tracker initialization within the OT module or for effective QRL.}
    \label{fig:small_boxes}
\end{figure}

\paragraph{Data cleaning for EgoTracks.}
Since EgoTracks contains annotations in contiguous frames, we uniformly sampled one-third of all frames. EgoTracks contains annotations for very small objects to assess the ability of visual object trackers to continuously follow objects subject to scale changes. Since these objects would be unfeasible to detect from single frames and may be source of ambiguity due to occlusion or motion blur, we discarded bounding box annotations with areas falling in the first quartile of all bounding box areas in the training set. Figure \ref{fig:small_boxes} shows a few examples of annotation bounding boxes that fall in the first quartile and therefore discarded from the dataset.
This leads to a reduced training set of $474,761$ annotated images, while we consider a validation set of $2,466$ images.

\paragraph{Faster R-CNN - based detector.} A first version of the object discovery module is obtained by training a Faster R-CNN instance on the above described dataset with object class taxonomies obtained through clustering of object descriptions. The performance of this module is limited as shown later in the quantitative evaluation, so this implementation of the OD module is not considered in the main paper.

\paragraph{YOLO - based detector.}
A second version of the object discovery module was based on the YOLOv10~\cite{wang2405yolov10} object detector, given its capacity for real-time inference. 
We trained the YOLOv10 detector on the same set of refined EgoTracks annotations.
Given the sparsity of EgoTracks annotations, both the Faster R-CNN and YOLO detectors tend to recognize a few object instances per image, which may limit the ability of the object discovery module to identify all possible objects of interest.
To address this issue, we trained the YOLOv10 object detector on the EgoObjects~\cite{zhu2023egoobjects} dataset, which contains dense object annotations for egocentric frames with a fixed taxonomy of $638$ classes. 
To make the EgoObjects and EgoTracks sets compatible, we mapped each EgoTracks object description to its corresponding class from the EgoObjects taxonomy. This is done by first describing the object description and class name with a \textit{Sentence Transformer}, and then selecting the class maximizing the cosine similarity between the embeddings. To reduce noise, we discarded all annotations with a similarity lower than $0.7$.

\begin{table*}[t]

\centering
\fontsize{9}{8}\selectfont
\begin{tabular}{@{}l | llllll@{}}
\toprule
EgoObjects Classes &
  accordion &
  air conditioner &
  headphones &
  washing machine &
  ... &
  yoga mat \\ 
  \midrule
\multicolumn{1}{l|}{EgoTracks Mapped Classes} &
  \multicolumn{1}{c}{-} &
  \multicolumn{1}{l}{\begin{tabular}[c]{@{}l@{}}air conditioner, \\ air conditioner  control, \\ air conditioner., \\ air conditioning machine\end{tabular}} &
  \multicolumn{1}{l}{\begin{tabular}[c]{@{}l@{}}head phones, \\ headphone, \\ headphones, \\ headset, \\ earphone, \\ earphones\end{tabular}} &
  \multicolumn{1}{l}{\begin{tabular}[c]{@{}l@{}}flat washer, \\ washing machine\end{tabular}} &
  \multicolumn{1}{l}{...} &
  \multicolumn{1}{l}{\begin{tabular}[c]{@{}l@{}}gym mat, \\ mat, \\ workout mat, \\ yoga mat, \\ exercise mat, \\ fitness mat\end{tabular}} \\ \bottomrule
\end{tabular}
\caption{\textbf{Examples of class mappings from EgoObject classes to EgoTracks object descriptions.} The first row reports EgoObject classes, while the second row contains descriptions of EgoTracks matched with embedding similarity.}
\label{tab:classes-mapping}
\end{table*}

\paragraph{Examples of class mapping.}
Table \ref{tab:classes-mapping} shows some examples of object descriptions retrieved from the EgoTracks annotations and mapped to EgoObjects objects. Note that some objects such as ``air conditioner", ``headphones" or ``yoga mat" have a reasonable match, while ``washing machine" contains ``flat washer", which has a different meaning despite a close embedding. We did not manually adjust these inconsistencies to keep the process fully automated. The table contains one class with no matches (accordion), due to the absence of EgoTracks objects with similar descriptions. Out of the total number of class descriptions (3491), we mapped 1,241 captions, while the remaining 2,250 classes were excluded from the dataset. This resulted in a high-quality subset of the data suitable for training and evaluation.

\paragraph{Quantitative comparison.}
Despite the domain difference between EgoObjects and EgoTracks, we noticed improved qualitative and quantitative results.
To further adapt the model to the EgoTracks domain, we fine-tuned the YOLOv10 object detector, pre-trained on EgoObjects, on the EgoTracks dataset with labels mapped to the EgoObjects taxonomy.
Table \ref{obj_det_metrics} compares the performance of the considered approaches with three off-the-shelf object detectors: GroundingDino~\cite{liu2023grounding}, the Hand-Object-Detector of~\cite{shan2020understanding}, and YoloWorld~\cite{cheng2024yolo}. We do not train these detectors on the EgoTracks data and use the official checkpoints and implementations provided by the authors. Since Visual Query Localization does not require to correctly identify object classes, all methods are evaluated with AP and Recall, merging all ground truth and predicted classes in a single ``object'' class. 
It is worth noting that AP and Recall measure different aspects of the detector. Indeed, AP measures the average ratio between correct predictions and all predictions made by the detector, whereas Recall measures the fraction of objects which have been correctly found by the detector, regardless of the total number of objects. In our settings, it is desirable to have a large recall (future queries have more chance to end up in the memory), but, since it is not possible to track all objects and a selection must be made based on the confidence score, in practice, a very low value of AP is not desirable. The harmonic mean between AP and Recall is hence adopted as a way to identify the most balanced detector. 
Results in Table \ref{obj_det_metrics}, obtained with a confidence score threshold $\lambda_{od} = 0.01$, show that Faster R-CNN trained on labels clustered from EgoTracks achieves the second-highest recall ($0.5224$), but also yields a very low AP value ($0.0219$). On the contrary, the YOLOv10 instance achieves the highest AP ($0.1983$) but also a low recall ($0.1634$). We attribute these skewed results to the sparse nature of annotations in EgoTracks (only a few object instances annotated per frame) and the different ways the two methods handle negative examples. The harmonic mean results of the first one are rather low ($0.0420$), while the second result is higher but not optimal at $0.1791$, highlighting the unbalanced nature of the two approaches. Training YOLOv10 on EgoObjects, which contains much denser annotations increases Recall to $0.3780$ at the expense of AP ($0.0247$), which leads to a low harmonic mean of $0.0463$. Mapping EgoTracks annotations to the EgoObjects taxonomy and training YOLOv10 using both datasets provides the most balanced detector, with a harmonic mean of $0.1882$, an AP of $0.1358$ and a Recall of $0.3067$. 

YOLOWorld achieves the lowest precision value among all the methods (0.0207), but the highest recall, equal to 0.5969. Due to the low precision value, the harmonic mean has a low score equal to 0.0401.
GroundingDino achieves a low AP and a reasonable recall value ($0.0450$ and $0.2572$), which is likely due to the large domain shift between data used for training GroundingDino and the challenging egocentric observations of EgoTracks. Using a Hand-Object-Detector as an object detector leads to lower results, which is because the Hand-Object-Detector only focuses on interacted objects, hence neglecting many non-interacted objects which may nevertheless be queried in the future.
Lastly, RT-DETR (used in VideoAgent \cite{fan2025videoagent}) achieves results comparable to GroundingDino, due to similar reasons.

\paragraph{Qualitative Examples.}
Figure \ref{fig:obj_det_pred} reports qualitative examples, reporting the predictions of the discussed models on a frame from the validation set of EgoTracks. Results are consistent with the evaluation metrics in Table \ref{obj_det_metrics}, Faster R-CNN (A) is the model with the second-highest recall, indeed it is the model with the highest number of bounding boxes containing both correctly localized objects and errors. YOLOv10 trained on EgoTracks (B) has fewer predictions than Faster R-CNN, but they are more accurate. Training YOLOv10 on EgoTracks (C) shows similar results to Faster R-CNN, with the second-densest predictions, translated into the second-highest recall model. YOLOv10 trained on EgoObjects fine-tuned in EgoTracks (D) has the most balanced results; predictions are well localized, reflecting the best-performing model in the HM metric. 
YOLOWorld (E) achieves results comparable to Faster R-CNN (A), demonstrating the highest recall with dense object predictions and minimal errors, which can be further reduced by increasing the threshold. 
i%
GroundingDino (F) contains several correctly regressed objects, but as shown in the frame, no ground truth annotations were found. Hand-Object Detector (G) has the worst performance, containing only two predictions, one of which is related to the interacted object. The discovery mechanism of this detector leads to the lowest values for both precision and recall. The predictions of VideoAgent (H) are partially correct, despite the presence of a few wrong bounding boxes. Finally the predictions marked with I are the ground truth ones.

\paragraph{OD$_{yolo}$ and OD$_{gdino}$.}
Given these results, in the following experiments, we consider the YOLOv10 detector trained on EgoObjects and EgoTracks as a representative of trained methods and both GroundingDino and Hand-Object-Detector as representative of off-the-shelf methods. These are referred to as OD$_{yolo}$ and OD$_{gdino}$ respectively in the main paper.

\paragraph{OD$_{oracle}$}. The oracular object detector is obtained using the ground truth object annotations in EgoTracks as valid detections. As previously mentioned, EgoTracks is aligned to the Episodic Memory benchmark of Ego4D, so EgoTracks detections include valid queries.

\label{sec:supp_OT}
\begin{table}[t]
\tblalternaterowcolors
    \centering
    \resizebox{\linewidth}{!}{
    \begin{tabular}{l | l | cc c}
    \toprule
    Detector & Training Dataset & \ap & \rec\  & HM $\uparrow$ \\
    \midrule
    Faster R-CNN & EgoTracks$^*$ & 0.0219& 0.5224 &0.0420\\
    YOLOv10 & EgoTracks$^*$ & \textbf{0.1983}& 0.1634 &0.1791\\
    YOLOv10 & EgoObjects & 0.0247& 0.3780 &0.0463\\
    YOLOv10 & EgoObjects + EgoTracks$^{**}$ & 0.1358 & 0.3067 &\textbf{0.1882}\\
    \midrule
    YOLOWorld & - & 0.0207 & \textbf{0.5969} & 0.0401\\
    
    GroundingDino & - & 0.0450 & 0.2572 &0.0765\\
    Hand-Object Detector& - & 0.0247 & 0.0381 & 0.0299\\
    
    \midrule
 RT-DETR & - & 0.0418& 0.3374 &0.0743\\
 \bottomrule
    \end{tabular}
    }
    \caption{\textbf{Comparison of the different object detectors.} $^*$ denotes that the object taxonomy has been obtained through clustering. $^{**}$ denotes that classes have been obtained by mapping EgoTracks descriptions to the EgoObject taxonomy. 
    HM: Harmonic Mean between AP and Recall. (Best results are highlighted in \textbf{bold}.)}
    \label{obj_det_metrics}
\end{table}

\begin{figure}[t]
    \centering
    \includegraphics[width=\columnwidth]{images/obj_det_qualitative.pdf}
    
    \caption{\textbf{Visualization of predictions from the object detection models.} Here we qualitatively show the performance of the different object detectors over the same sampled frame.
    A: Faster R-CNN on EgoTracks, B: YOLOv10 on EgoTracks, C: YOLOv10 on EgoObjects, D: YOLOv10 on EgoObjects, fine-tuned in EgoTracks, E: YOLOWorld, F: GroundingDino, G: is the Hand-Object Detector H: RT-DETR, I: Ground-truth annotations.}
    \label{fig:obj_det_pred}
\end{figure}

\subsection{Query Retrieval and Localization (QRL)}
\label{QRL}
\label{sec:supp_qrl}
We implemented two distinct approaches for localizing a visual query $\query$ within a memory $\memory$: an implementation based on SiamRCNN with a classification head \cite{voigtlaender2020siam}(QRL$_{siamrcnn}$); and an approach based on DINOv2 with cosine similarity \cite{goletto2024amego,oquab2023dinov2} (QRL$_{dino}$). \\
In QRL$_{siamrcnn}$, a Feature Pyramid Network (FPN) \cite{lin2017feature} serves as the backbone $\Phi(\cdot)$ for features extraction, generating features for the visual representation of the object $\Phi(\phi_{i,t}),  \phi_{i,t} \neq \emptyset$ stored in $\memory$ as well as the features for the visual query $\Phi(\query$).  To determine whether $\query$ exists within $\memory$, each $\Phi(\phi_{i,t})$ is compared with $\Phi(\query$) using a siamese network head, $\Psi(\Phi(\phi_{i,t}),\Phi(\query))$, which implements a bilinear operation \cite{voigtlaender2020siam} that predicts instance similarity score belonging to the interval [0, 1].

In QRL$_{dino}$, DINOv2 \cite{oquab2023dinov2} is employed as the backbone $\Phi(\cdot)$ for extracting feature representations $\Phi(\phi_{i,t})$ and $\Phi(\query)$. In this approach, the similarity operation $\Psi(\Phi(\phi_{i,t}),\Phi(\query))$ is implemented using cosine similarity \cite{goletto2024amego}, which provides scores belonging to the interval [0, 1].\\

In both cases, for each object i, the similarity scores are
averaged into a single score $r_i$. If the maximum score $r_i$
exceeds the threshold $\lambda_{ref}$ = 0.5, the object $\object_i$ is considered a match for query $\query$. Once a match is identified, the most recent sequence of
contiguous bounding boxes from $\object_i$’s spatio-temporal history is retrieved and returned as the response track \textbf{r}. \\

\subsection{ESOM Memory Overhead}
The GPU memory overhead is 423MB for OD$_{yolo}$ and 2,350MB for OT$_{egostark}$ when tracking $10$ objects concurrently, 2,320MB for OT$_{masa}$, and 2,474MB for QRL$_{siamrcnn}$.

\section{Online Object Memory Baselines}
\label{sec:baselines}
Here we provide details of the implementations of the baselines used for comparison.

\subsection{AMEGO}
\label{sec:supp_amego}
The AMEGO framework \cite{goletto2024amego}, was employed to construct an episodic memory model, serving as a performance benchmark for our proposed approach. We utilized \textit{ce683bd} version of the AMEGO codebase, available at \footnote{\texttt{\href{https://github.com/gabrielegoletto/AMEGO.git}{https://github.com/gabrielegoletto/AMEGO.git}}}, with minor modifications to accommodate the EGO4D dataset structure. Our implementation focused specifically on the Hand-Object Interaction (HOI) module of AMEGO. The Location Segmentation module was excluded from our analysis as it fell outside the scope of our research objectives. To ensure a fair and consistent comparison, we maintained the original hyperparameters and thresholds to align with the baseline model published on the official GitHub repository. Since their experiment has been made using a different dataset, our modifications were limited to data input/output functions to handle the EGO4D data format, without altering the core algorithmic components.

\subsection{VideoAgent}
\label{sec:supp_videoagent}
The VideoAgent framework \cite{fan2025videoagent}, has been utilized as a comparative baseline for our episodic memory model. We employed the \textit{21e4eb7} version of the VideoAgent repository\footnote{\texttt{\href{https://github.com/YueFan1014/VideoAgent.git}{https://github.com/YueFan1014/VideoAgent.git}}}.
Our implementation focused on the Object Memory module, and we opted to omit the captioning module since it was not pertinent to our research scope. To maintain experimental integrity, we preserved the original configuration parameters.

\subsection{Memory Representation} As with the other OMP configurations discussed in this paper, the object tracklets produced by VideoAgent and Amego OMP are stored in memory using the MR1* representation, as it provides the best balance between accuracy and memory efficiency (see Table \ref{tab:memoryrep}). Specifically, for each $\object_i$ found by either AMEGO or VideoAgent object detection and tracking algorithms, $\memory$ retains the RGB patch ($\phi_{i,t}$) and the $\bbox_{i,t}$ at the time of object discovery, along with two RGB patches from the most recent response track. These patches correspond to the first and last valid patches identified by the classifier within the most recent sequence of bounding boxes.  
For all other temporal locations not covered above, $\memory$ stores only the spatio-temporal information provided by $\bbox_{i,t}$.

\subsection{Query Retrieval and Localization (QRL)} To localize the visual queries $\query$ within the memory $\memory$ built over AMEGO and VideoAgent, we used the implementation based on SiamRCNN with a classification head (QRL$_{siamrcnn}$). Once the similarity scores are computed for the different patches, they are averaged into a single score $r_i$. If the maximum score exceeds the $\lambda_{ret}$ = 0.5, the object $\object_i$ is considered a match for query $\query$.

\section{Evaluation Dataset}
\label{sec:supp_dataset}
Due to the high storage requirements for the validation split of the Ego4D VQ2D benchmark \cite{ego4d} and the significant time needed for experiments on the full dataset, we opted to use a subset of the validation clips for our evaluations. Specifically, we randomly selected 115 video clips uniformly, containing 450 visual queries along with their corresponding ground-truth response tracks and tracklets. This subset consisted of a total of 208,365 frames. To maintain consistency with the original frame rate of the EGO4D dataset \cite{ego4d}, we ran the OMP algorithm at 5 frames per second.

\begin{table}[t]
\fontsize{7}{7}\selectfont
	\centering

    \setlength\tabcolsep{.1cm}

\begin{tabular}{ l l |c c c|c} 

\toprule
\multicolumn{2}{l|}{OMP Configuration} & \multirow{2}{*}{\tAPZS} & \multirow{2}{*}{\stAPZS} & \multirow{2}{*}{\succ} & \multirow{2}{*}{\size} \\
 OD & OT  &  &  &  &  \\

\midrule

 \rowcolor{tblrowcolor2}  OD$_{oracle}$ & {OT$_{egostark}$}  & \textbf{18.3}  & \textbf{5.1} & \textbf{31.91} & \textbf{591.4 MB}\\

 OD$_{yolo}$ & OT$_{egostark}$  & 0.03  & 0.2 & 4.02 & 1.7 GB MB\\ 

\midrule

\rowcolor{tblrowcolor2}  OD$_{oracle + yolo}$ & OT$_{egostark}$  &  10.6 & 2.3 & 23.67 & 2.4 GB\\

\bottomrule		
\end{tabular}

\caption{\textbf{More objects in the memory can confound QRL.} In this table, by combining the capabilities of the OD$_{oracle}$ and OD$_{yolo}$, we tested wether the QRL method is robust to a memory $\memory$ which contains the true visual queries as well as other object found while processing $\video$. (Best results are in \textbf{bold}).
} 
\label{tab:additionalOMP}
\end{table}

\section{Additional Results}
\textbf{Combination of OD modules.} The performance of ESOM has been evaluated under various configurations of the OD and OT modules (Table  \ref{tab:discoverytracking}). However, OMP configurations using OD$_{oracle}$ construct a memory that includes only objects that will be queried. To incorporate additional legitimate objects into $\memory$, we combined the oracle detections, together with the OD$_{yolo}$detections (see OD$_{oracle+yolo}$ in Table \ref{tab:additionalOMP}). Adding more objects to memory reduces ESOM's performance compared to the OMP configuration with OD$_{oracle}$. This suggests that an increased number of objects in memory hinders the QRL algorithm's ability to accurately discriminate the visual query, resulting in confusion.
However, its performance remains higher than the OMP configuration using only OD$_{yolo}$, further highlighting the challenges detectors face in identifying objects that will appear as queries in the future.

\textbf{Random performance}
The performance of ESOM has been evaluated by implementing a QRL method that randomly (w. uniform dist.) selects a response track from $\memory$. This achieves a Succ. score of 27.29 (OD$_{oracle}$+OT$_{oracle}$) and of 0.44 (OD$_{yolo}$+OT$_{egostark}$), while ESOM achieves 81.92 and 4.02 respectively. ESOM performance, even in suboptimal OMP configuration (OD$_{yolo}$+OT$_{egostark}$) are still better than random selection. This indicates that the method effectively leverages available information to enhance performance.

\textbf{Constant memory budget}
We explored a “constant-budget” version of ESOM, in which older and low-confidence objects are pruned as soon as a predefined memory limit is reached. Results (Figure \ref{fig:success_and_time_comparison}) demonstrate that ESOM isn't particularly affected by memory constraints.

\begin{figure}[t]
    \centering
\includegraphics[width=\linewidth]{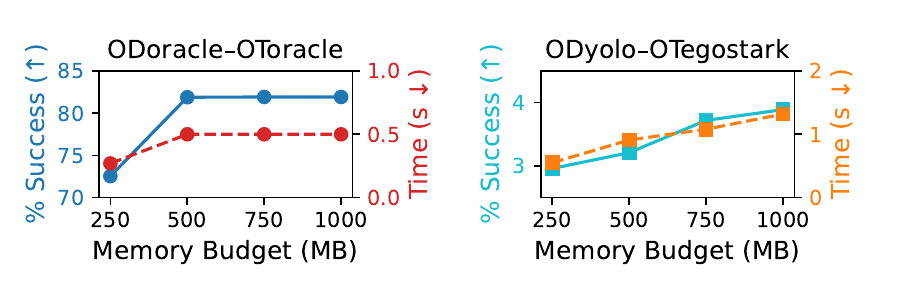}
    \caption{\textbf{Impact of memory-budget on ESOM performances.} 
   Success rate and query time for two ESOM variants under varying memory budgets (250–1000 MB). Results show that performance remains stable as older or low-confidence objects are pruned to meet memory constraints, indicating ESOM's robustness under limited memory conditions.}
    \label{fig:success_and_time_comparison}
\end{figure}

\paragraph{Qualitative examples.}
\label{sec:supp_qualitative}
Figure \ref{fig:qualexsupp} shows additional qualitative examples of the behavior of QRL in retrieving and localizing different visual queries within two memory representations build by two OMP configurations.

\begin{figure}[t]
    \centering
\includegraphics[width=\columnwidth]{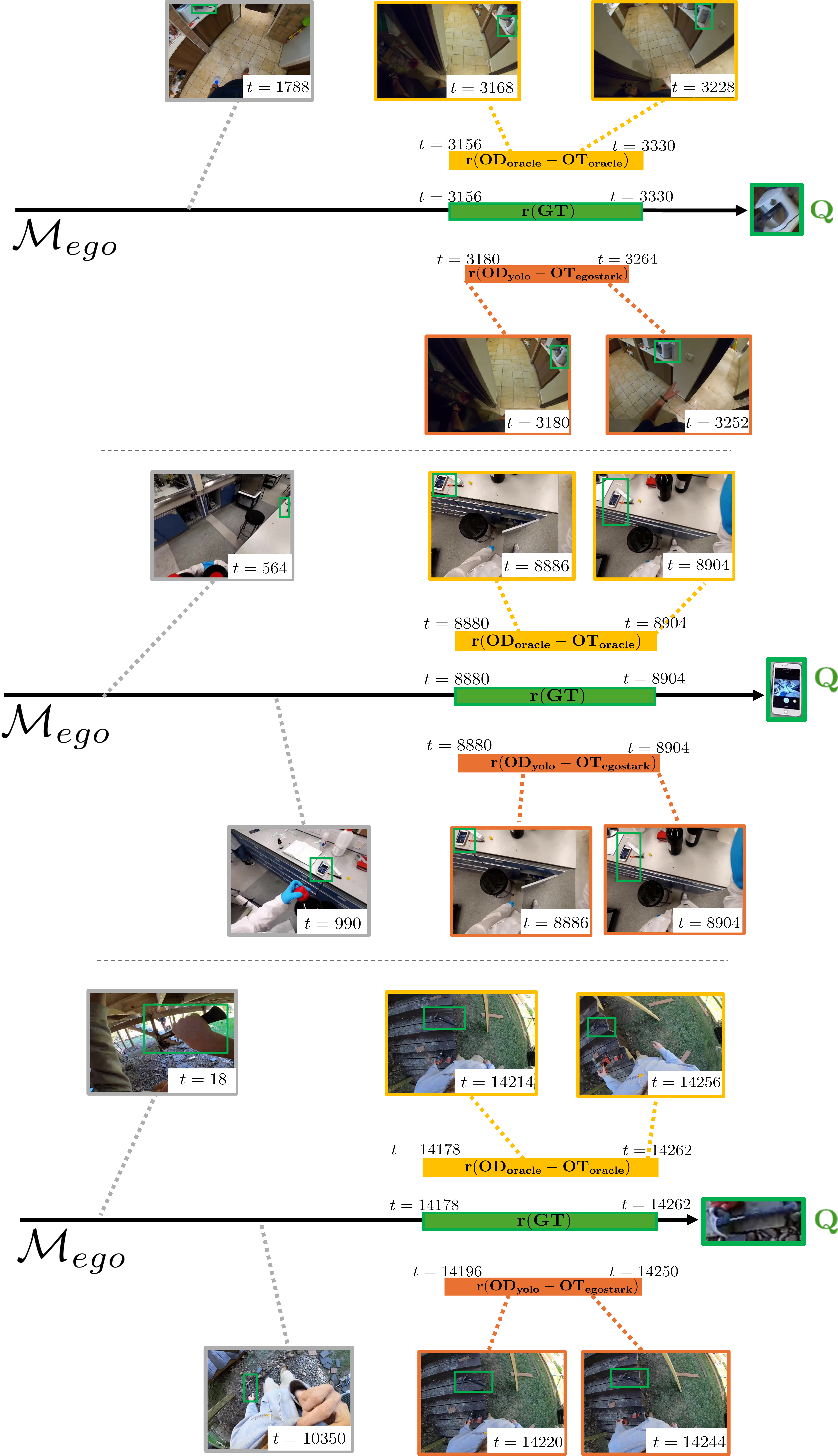}
    \caption{\textbf{Additional qualitative examples.} 
    Successful retrieval of thre different queries \textcolor[HTML]{4DA72E}{$\query$} from {$\memory$} using the oracle OMP algorithm (\textcolor[HTML]{CC9900}{OD$_{oracle}$ - OT$_{oracle}$}) and the real-world OMP (\textcolor[HTML]{ED7D31}{OD$_{yolo}$ - OT$_{egostark}$}). 
Sample frames from each track are shown with corresponding \textcolor[HTML]{4DA72E}{bounding boxes} whose patches $\phi_{i,t}$ were matched during retrieval. The frames in \textcolor[HTML]{808080}{gray} correspond to the moment of the object's discovery. The \textcolor[HTML]{4DA72E}{green bar} visualizes the temporal extent of the ground-truth track, with timestamps indicating its duration.}
    \label{fig:qualexsupp}
\end{figure}

\end{document}